\documentclass[11pt]{article}
\PassOptionsToPackage{table}{xcolor}
\PassOptionsToPackage{numbers,square,sort&compress}{natbib}
\usepackage{natbib}
\usepackage{IEEEtrantools}
\let\BARthebibliography\thebibliography
\let\BARendthebibliography\endthebibliography
\let\BARbibliographystyle\bibliographystyle
\renewcommand{\bibliographystyle}[1]{}
\let\BARsavedaddcontentsline\addcontentsline
\usepackage[preprint,nohyperref]{acl}
\let\bibliographystyle\BARbibliographystyle
\let\thebibliography\BARthebibliography
\let\endthebibliography\BARendthebibliography
\let\addcontentsline\BARsavedaddcontentsline
\usepackage[T1]{fontenc}
\usepackage[utf8]{inputenc}
\usepackage{mathptmx,helvet}

\usepackage{microtype,etoolbox,amsmath,amssymb,booktabs,tabularx,array,longtable}
\usepackage{xcolor}
\definecolor{barText}{HTML}{333333}
\definecolor{barBlue}{HTML}{0072B2}
\definecolor{barOrange}{HTML}{D55E00}
\definecolor{barOrangeText}{HTML}{B34F00}
\definecolor{barRose}{HTML}{AD5868}
\definecolor{barGreen}{HTML}{4D7864}
\newcommand{\auditerror}[1]{\textcolor{barRose}{#1}}
\newcommand{\auditcorrection}[1]{\textcolor{barGreen}{#1}}
\usepackage{caption}
\usepackage[most]{tcolorbox}
\usetikzlibrary{arrows.meta}
\usepackage{placeins,float}
\usepackage{needspace}
\usepackage{etoc}
\usepackage{xurl}
\let\cite\citep
\usepackage[colorlinks=true,linkcolor=barText,citecolor=barText,urlcolor=barText]{hyperref}
\hypersetup{pdftitle={Follow the Policy, Fail the Benchmark: Auditing and Revising Policy-Based Decision Benchmarks},
pdfauthor={Yanru Xiao and Yifei Cai}}
\usepackage{enumitem}
\setlist{nosep,leftmargin=*}

\newcommand{\tablefont}{\small}
\newcolumntype{L}[1]{>{\raggedright\arraybackslash}p{#1}}
\newcolumntype{Y}{>{\raggedright\arraybackslash}X}
\newcommand{\artifact}[1]{\nolinkurl{#1}}
\newcounter{auditcase}
\newenvironment{auditcase}[3][]{%
  \refstepcounter{auditcase}%
  \begin{tcolorbox}[enhanced,breakable,
    colback=white,colframe=barBlue!65,colbacktitle=barBlue,coltitle=white,
    boxrule=.45pt,arc=1mm,left=2.5mm,right=2.5mm,top=1.5mm,bottom=1.5mm,
    before skip=6pt,after skip=6pt,
    fonttitle=\small\sffamily\bfseries,fontupper=\small,
    before upper={\setlength{\parskip}{2pt}%
      \interlinepenalty=10000\clubpenalty=10000\widowpenalty=10000},
    title={Case \theauditcase\quad #2},
    title after break={Case \theauditcase\ (continued)},#1]
  \label{#3}%
}{\end{tcolorbox}}
\newcommand{\casesource}[1]{%
  {\footnotesize\sffamily\bfseries\color{barBlue} Source material.}\
  {\footnotesize\sffamily #1}\par}
\newcommand{\caseaudit}[1]{%
  \par\smallskip\noindent\vbox{%
  \noindent\textcolor{barOrange!65}{\rule{\linewidth}{.4pt}}\par\smallskip
  {\footnotesize\sffamily\bfseries\color{barOrangeText} Our audit.}\
  {\footnotesize\sffamily #1}\par}\par\nopagebreak[4]}
\newcommand{\casefield}[2]{\par\textbf{#1.} #2\par}
\title{Follow the Policy, Fail the Benchmark:\\
Auditing and Revising Policy-Based Decision Benchmarks}
\author{Yanru Xiao\\
Amazon Web Services\\
\texttt{sugarruy@gmail.com}
\And
Yifei Cai\\
Iowa State University\\
\texttt{yifeic@iastate.edu}}
\date{}
\begin{document}
\bstctlcite{BAR:bibcontrol}
\etocdepthtag.toc{main}
\maketitle
\begin{abstract}

Policy-based decision benchmarks evaluate whether systems apply written rules to financial and operational cases. Yet a rule-following response can be marked wrong when the benchmark misstates the required decision. 
We audit nine public policy-based decision benchmarks through source tracing, program execution, and controlled checks. We find unsupported targets, reference errors that survive replay or propagate as solved steps, and evaluators that reject acceptable answers or credit incomplete decisions.
To measure the scoring impact, we rescore unchanged model responses. On the 96 LegalBench SARA numeric cases, correcting the scorer raises acceptance by more than 80 percentage points for each of three models, without changing references or responses. In RuleArena Tax, restoring a required zero floor changes 27 of 300 references and narrows an observed model gap. A controlled federal-exemption task with explicit ownership and eligibility assumptions also shows that errors in supplied intermediate answers can affect final decisions. We present BAR (Benchmark Audit and Revision)\footnote{Code and revised data: \url{https://github.com/SugarRuy/BAR-Policy-Decision}.}, a versioned collection of audit evidence and revisions for reevaluation.

\end{abstract}

\section{Introduction}

When organizations apply written policies to individual cases, the resulting decisions can determine charges, refunds, access to services, or permission to act \citep{rulearena,taubench}. Public benchmarks use such tasks to compare direct prompting, translation into code or logic, and tool-using agents \citep{nair,codeprompt,taubench}. For meaningful comparisons, higher scores should reflect decisions better supported by the supplied policies and case facts.

\begin{figure*}[t]
\centering
\includegraphics[width=\textwidth]{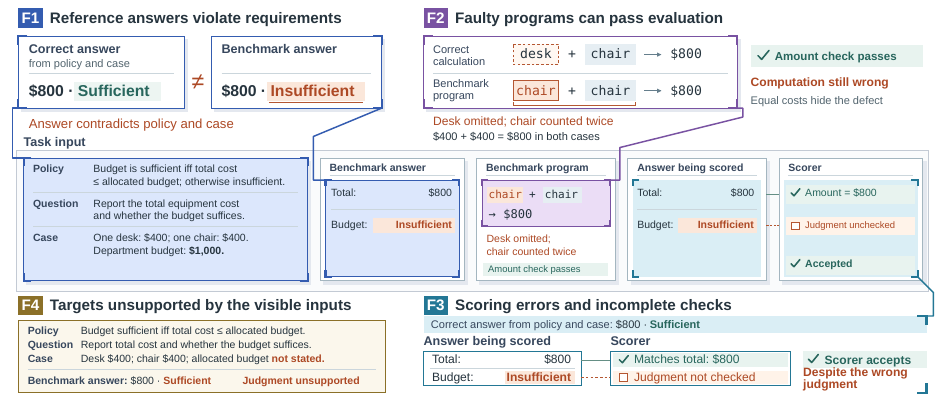}
\caption{Schematic illustration of the four classes of findings through independent
variants of a hypothetical equipment-budget task.
Matching colors connect selected materials to enlarged inspections; the separate case below illustrates missing information.
\hyperref[sec:mech-reference]{F1}: a benchmark answer contradicts the policy and case.
\hyperref[sec:deontic]{F2}: an incorrect benchmark program passes an amount check.
\hyperref[sec:mech-scoring]{F3}: a scorer accepts the correct amount with the wrong budget judgment.
\hyperref[sec:mech-targets]{F4}: a target is unsupported when the allocated budget is omitted.
The panels do not form an execution sequence.}
\label{fig:audit-overview}
\end{figure*}

However, even a response that follows an explicit rule can be marked wrong. In RuleArena Tax \citep{rulearena}, the released reference program omits a required zero floor, causing an answer that applies this floor to be rejected. Applying the floor changes the grade of the same response from rejected to accepted (Case~\ref{fig:opening-witness}). The reverse can also occur: an evaluator may accept a response that omits a required decision. In both cases, the reported score reflects not only model behavior but also defects in the benchmark. These problems can extend beyond evaluation. When a benchmark supplies reference answers as solved steps in later tasks, errors in those answers may influence subsequent model decisions. Figure~\ref{fig:audit-overview} illustrates the four kinds of failure using a constructed equipment-budget task.

\begin{figure}[!t]
\begin{auditcase}[unbreakable,arc=.5mm]{RuleArena Tax: a required zero floor}{fig:opening-witness}
\casesource{The task asks for tax owed (positive) or a refund (negative). In \texttt{tax/comp\_0/17}, line~22 states: ``Subtract line 21 from line 18. If zero or less, enter -0-''. Here, \(7{,}102-8{,}192=-1{,}090\).}
\caseaudit{Only this reference step changes; all other operations and defaults remain fixed.}
\smallskip
{\centering
\begin{tabularx}{\linewidth}{@{}Xrr@{}}
\toprule
Quantity & Original & Revised\\
\midrule
Line 22 & \(\auditerror{-1{,}090}\) & \(\auditcorrection{0}\)\\
Other taxes \( - \) payments & \(703\) & \(703\)\\
Final amount & \(\auditerror{-387}\) & \(\auditcorrection{703}\)\\
\midrule
Grade for answer \$703 & Rejected & Accepted\\
\bottomrule
\end{tabularx}\par}
\smallskip
This changes 27/300 references. Both versions still use zero for the missing Schedule~3 input.
\end{auditcase}
\end{figure}

Prior work has documented label errors and broader weaknesses in benchmark evaluation \citep{northcutt2021,benchguard2026}. We investigate these problems in policy-based decision benchmarks, where answers must be justified by the supplied policy, question, and case. Specifically, we examine how benchmark defects affect the evaluation of unchanged model responses and whether faulty intermediate answers influence subsequent decisions. Our contributions are:

\begin{enumerate}
\item A source-grounded audit of nine public policy-based decision benchmarks. We identify defects in inputs, references, programs, and evaluators, including errors propagated into later tasks and program faults concealed by accepted outputs.

\item Measurements through fixed-output rescoring and an experiment with original, revised, or omitted intermediate answers.

\item BAR (Benchmark Audit and Revision), a versioned collection of audit evidence and scoped revisions.
 It distinguishes corrections that permit rescoring existing responses from changes requiring new predictions, while documenting conditional repairs and unresolved findings.
\end{enumerate}

\section{Task, selection, and audit standards}
\label{sec:scope}

A task supplies rules forming a normative or procedural \emph{policy} \(P\), a question \(Q\), and case information \(C\). The model returns an amount, categorical decision, conditional answer, or information request. The benchmark also provides a reference answer, optionally a program that computes it, and a scoring function. The target is the set of responses supported by these inputs. A single reference decision is justified when the policy and case determine it; conditional or discretionary tasks may admit several decisions. We assess underdetermination using all model-visible information, including demonstrations, and any permitted external knowledge. For interactive tasks, we check the final action against terminal rules, given the agent's observations.

The policy and case must have an identifiable source and supply information
relevant to the decision; they need not occupy fields named \(P\) and \(C\).
A question may be standardized for evaluation. Changing the requested
decision, however, requires a compatible target and defines an adaptation.
When attributing discrepancies, we distinguish original releases, published
derivatives, and our experimental inputs.

\subsection{Selection and source lineage}
\label{sec:selection}

We purposively select tasks that require applying a supplied policy to a question and case, and whose inputs, code, or scorers are available for inspection. We cover RuleArena \citep{rulearena}, OpenExempt \citep{openexempt}, LegalBench SARA \citep{legalbench}, DeonticBench \citep{deonticbench}, ConditionalQA \citep{conditionalqa}, ShARC \citep{sharc}, PolicyGuardBench \citep{policyguardbench}, WorkArena++ \citep{workarena}, and TheAgentCompany \citep{theagentcompany}.

LegalBench SARA and DeonticBench share SARA material \citep{sara} but count separately because their inputs, programs, and scorers differ. Policy-as-Logic \citep{nair} evaluates all three RuleArena domains and Code Prompting \citep{codeprompt} reports ConditionalQA. Reuse alone does not establish use of the audited component or version, so we attribute no effect to those published scores. OpenExempt is included for its inspectable generation code.

The selection covers monetary calculations, conditional answers, clarification requests, trajectory judgments, and actions in an environment. DeonticBench provides reference programs for checking internal computations; WorkArena++ and TheAgentCompany provide evaluators for checking completed actions. Appendix~\ref{sec:task-contracts} gives the task contracts and verified reuse. Historical banking repairs and SOP-Bench decision-rule controls supplement the nine benchmarks without adding to that count.

\subsection{Audit evidence and correction criteria}
\label{sec:standards}

We check each reported finding against its frozen source release. A finding identifies the affected component (policy, visible input, reference target or program, or evaluator) and source version. It provides a complete \emph{witness} (case, source passage, or execution trace), a discriminating control or source trace, and the evaluation consequence. Our checks use recorded executions, grouping of identical model-visible inputs, tests of scoring conditions on constructed inputs, and model-assisted semantic readings. These readings consider applicability, priorities, exceptions, and definitions.

Coverage ranges from all-case structural checks to selected witnesses. Structural counts describe their specified releases; selected semantic findings do not estimate benchmark-wide error prevalence. Replacement targets must follow from visible inputs and remain conditional on any source annotations used; otherwise the target is unresolved or the affected record is explicitly excluded. Appendix~\ref{sec:versions} gives the inspected populations, review procedure, negative checks, and remaining uncertainties.

\section{Findings}
\label{sec:findings}

We identify failures in reference answers and programs, scoring procedures, and the alignment between visible inputs and assigned targets. See \mbox{Appendix~\ref{sec:more-cases}} for case evidence and controls.

\subsection{Reference answers violate requirements}
\label{sec:mech-reference}

Reference answers used as ground truth can inherit errors from the programs that generate them. These errors can affect the final evaluation target or intermediate answers supplied to later tasks.

\paragraph{RuleArena Tax.} The reference answer in Case~\ref{fig:opening-witness} is inconsistent with the supplied zero-floor instruction. The final-answer program omits the floor, although the rule-level analyzer applies it. Holding all other operations and defaults fixed, we restore the line-22 floor and recompute downstream totals, changing 27 of 300 reference answers (Appendix~\ref{sec:tax}).

\paragraph{OpenExempt.}\label{sec:openexempt}
The valuation task asks for the maximum amount of each asset that can be protected under each exemption provision \citep{openexempt}. Its reference calculation applies each exemption only once. For some assets, the authors' annotations permit more under the same provision than the reported maximum.

These values are then supplied to models as solved steps in later tasks. The final allocation program can include the omitted amount and exceed the supplied maximum (Case~\ref{case:oe-downstream}). This conflicts with the intermediate answer; it does not establish that the final allocation is also wrong. Rerunning the same valuation code reproduces the published values, including the inconsistent ones (Appendix~\ref{sec:oe-recurrence}).

\begin{figure}[!t]
\begin{auditcase}[unbreakable,arc=.5mm,halign title=flush left]{OpenExempt: a final allocation \mbox{exceeds} the reported maximum}{case:oe-downstream}
\casesource{In optimal-exemption stage 4 (OE4), record 2, the task asks how much property can be protected from creditors. The paper defines the intermediate amount for each asset and exemption provision as the ``maximum protectable dollar value''. For the same \textbf{\$8,200 car} under the same vehicle-exemption provision, \S522(d)(2):}
\par\smallskip
{\centering
\begin{tabularx}{\linewidth}{@{}>{\raggedright\arraybackslash}Xr@{}}
\toprule
Reference artifact & Amount\\
\midrule
\textcolor{barBlue}{Intermediate amount supplied as a \mbox{solved step}} & \textcolor{barBlue}{\textbf{\$4,450}}\\
\textcolor{barOrangeText}{Final reference allocation} & \textcolor{barOrangeText}{\textbf{\$8,200}}\\
\bottomrule
\end{tabularx}\par}
\smallskip
\caseaudit{The final allocation exceeds the intermediate amount that the paper defines as the maximum. The intermediate calculation applies the provision once, allocating \$4,450. This leaves \(\$8{,}200-\$4{,}450=\$3{,}750\). The final calculation applies the same provision again to cover this remainder, protecting the full \$8,200 car.}
\smallskip
The comparison establishes a conflict between reference artifacts without independently certifying either amount as legally correct.
The source annotation permits two claims and encodes an \$8,900 combined allowance.
\par\smallskip
The conflict affects \textbf{38 asset--provision entries in 33 of the 100 records} in this final-allocation component.
\end{auditcase}
\end{figure}

\subsection{Faulty programs can pass evaluation}
\label{sec:deontic}

DeonticBench provides statutes and case facts and evaluates either direct
model answers or answers obtained by executing model-generated Prolog
\citep{deonticbench}. It also releases reference programs for analysis and
training. Some of these programs contain errors yet produce outputs accepted
by the authors' scorer. In a SARA tax case, two exemption errors cancel: the program counts a dependent twice and omits the
taxpayer's own exemption (Case~\ref{case:deontic-tax}).

Another reference program is asked whether a stated tax amount follows
from the facts and statute; its answer, yes, matches the released label.
Its bracket clauses omit lower income limits, however, so it also derives
two other amounts for the same return that the statute does not support
(Appendix Case~\ref{case:deontic-probes}).

\begin{figure}[!t]
\begin{auditcase}[unbreakable]{DeonticBench SARA: two errors preserve the exemption total}{case:deontic-tax}
\casesource{In released \artifact{tax_case_61}, Alice is the taxpayer and
her father Bob is a dependent under the represented eligibility facts.
The supplied \S151 grants \$2,000 for the taxpayer and each dependent.
The source rules treat unproved exceptions as absent; the qualification
facts and this scope are detailed in Appendix~\ref{sec:deontic-detail}.
The program derives Bob's dependency twice, sums \$4,000 in exemptions,
and produces an accepted tax answer.}
\caseaudit{The program counts proofs of dependency rather than distinct
dependents, and has no separate taxpayer allowance.}
\smallskip
{\centering
\begin{tabularx}{\linewidth}{@{}Xr@{}}
\toprule
Exemption calculation & Total\\
\midrule
Supported: (1 Alice + 1 Bob) $\times$ \$2,000 & \$4,000\\
Program: (0 Alice + 2 Bob proofs) $\times$ \$2,000 & \$4,000\\
\bottomrule
\end{tabularx}\par}
\smallskip
Removing only the duplicate proof source leaves one Bob allowance and
still omits Alice's. The exemption falls to \$2,000 and the tax changes.
This modified-program diagnostic exposes the compensation; it does not
complete a repair or establish a replacement label.
\end{auditcase}
\end{figure}

Programs reused as explanations, supervision, or policy engines need checks of their encoded facts and rules.

\FloatBarrier

\subsection{Scoring errors and incomplete checks}
\label{sec:mech-scoring}

\paragraph{Answer processing and score aggregation.} LegalBench's SARA numeric task asks for a tax amount, but its published scorer deletes decimal points before extracting an integer \citep{legalbench}. This can reject an answer within the scorer's tolerance (Case~\ref{case:lb-summary}).
The acceptance formula also rejects every integer when the reference is zero, so the four zero-valued references among the 96 test cases cannot receive credit (Appendix~\ref{sec:lb-scoring-diagnostics}). Our separate correction accepts an explicitly parsed exact zero. Both defects concern LegalBench's SARA scorer, leaving the original SARA tax labels unaffected.

\begin{figure}[!b]
\begin{auditcase}[unbreakable,arc=.5mm]{LegalBench SARA: deleting a decimal point changes the grade}{case:lb-summary}
\casesource{The tax task instructs: ``State the amount first.'' In \texttt{tax\_case\_25}, the reference answer is \(r=\$259{,}487\) and a model response starts \texttt{\$259,487.13}. LegalBench's scorer applies \texttt{.replace(".", "")} before extracting integer \(p\). It accepts the response if \mbox{\(\lvert p/(r+0.1)-1\rvert<0.1\)}.}
\caseaudit{We keep the decimal point; response, reference, and scoring rule stay fixed.}
\par\smallskip
{\centering
\begin{tabularx}{\linewidth}{@{}Xrl@{}}
\toprule
Parser & Amount \(p\) & Accepted\\
\midrule
Original & \texttt{\auditerror{25948713}} & No\\
Keep decimal & \texttt{\auditcorrection{259487.13}} & Yes\\
\bottomrule
\end{tabularx}\par}
\end{auditcase}
\end{figure}

ConditionalQA asks models to answer questions from policy documents and give any conditions on the answers; an empty answer list denotes an unanswerable question. Its official evaluator records per-question metrics only for questions with a prediction: a missing prediction ID silently removes the question from the denominator, whereas a wrong but present answer is penalized. Controlled tests also expose division by zero for empty scoring categories (Appendix~\ref{sec:cqa-scoring}). Our revised evaluator scores every reference question, counts a missing prediction ID as zero, and marks metrics for empty categories as undefined.

A missing prediction ID can result from an interface failure, such as invalid output under a JSON interface; such a failure is not a label defect. Complete prediction files, including those published with Code Prompting \citep{codeprompt}, score identically under both versions.

RuleArena NBA asks whether a transaction violates league rules and which operation and team are responsible. Its separate rule-attribution metric checks whether the response mentions the annotated rules. Some annotated identifiers are absent from the extraction vocabulary, so attribution recall counts them as missed rules regardless of final-verdict accuracy (\mbox{Appendix~\ref{sec:nba}}).

\paragraph{Checking task requirements.}\label{sec:decision-coverage}
Even when responses are read correctly, the scoring conditions may omit a required decision or apply the wrong constraint. In a TheAgentCompany budget task \citep{theagentcompany}, the agent must tell each manager the total equipment cost and whether the budget suffices. We test this requirement through the official scoring chain with controlled replies in the task's actual manager chats. Replies that omit or explicitly reverse the budget decisions receive full credit when they retain the expected total strings. Changing an expected total loses credit. The evaluator therefore checks the totals but not the required budget judgments (Appendix Case~\ref{case:tac-budget}). The replies and the fixed trajectory input are constructed; no agent generates them.

Another TheAgentCompany task counts matching password records without checking that they cover distinct users. Repeating one valid record passes the file check while omitting other users (Appendix Case~\ref{case:tac-users}).

A selected WorkArena++ task \citep{workarena} asks an agent to schedule
requests under a strict limit on the gap between them. Its evaluator uses
a looser limit and accepts a noncompliant schedule
(Case~\ref{case:workarena-gap}).

\begin{figure}[!t]
\begin{auditcase}[unbreakable]{WorkArena++: a one-hour requirement checked as 24 hours}{case:workarena-gap}
\casesource{The selected tight-priority scheduling task requires at most
one hour between consecutive requests. The L2 prompt states the limit;
the L3 task refers to the scheduling manual with the same requirement.
The native validator instead selects a 24-hour limit for this class.}
\caseaudit{Constructed-state tests of the native validator.}
\casefield{Constructed state}{Two requests have valid durations, fit the
time window, and do not overlap, but have a two-hour gap.}
\casefield{Results and controls}{The tight-priority class accepts this state and its one-hour-gap
control. A separate plain-tight control rejects a two-hour gap, with risk
and duration settings appropriate to that mode. The selected class therefore
accepts a schedule that violates its requested constraint. These are
native-method checks on supplied states, not live agent episodes; L3
additionally requires the completed private-ticket state.
Appendix~\ref{sec:workflow-grading} gives the source branch and the
differences between control states.}
\end{auditcase}
\end{figure}

\subsection{Targets unsupported by the visible inputs}
\label{sec:mech-targets}

A target is \emph{unsupported} when the visible policy, question, and case do not justify it. A decision is \emph{underdetermined} when they leave its outcome open. Neither establishes the opposite label.

LegalBench's SARA derivative preserves original answers while selecting policy snippets for entailment questions. In a selected witness, the full source supports the label, but the snippet omits the required tax-rate schedule. Neither the standard nor the Claude-specific input template supplied by LegalBench includes the missing rates in its examples (Appendix~\ref{sec:lb-context}). Under a supplied-policy-only reading, success may require information beyond the input, and failure cannot isolate application of the missing rule. Restoring the rule creates a new input version that needs new predictions.

PolicyGuardBench supplies a policy and a fixed trajectory summary for binary
violation judgment \citep{policyguardbench}. Pooling train and test in each
release yields 279 groups of identical complete inputs with conflicting
binary labels, involving 780 records. The strict release removes train--test
trajectory overlap but preserves the labeled multiset and all these
conflicts (Appendix~\ref{sec:pgb}). These comparisons use the released
\texttt{instruction} and \texttt{input} fields; their exact binding to the
published evaluation harness remains unresolved.

The conflict remains within the held-out population: the strict test split
has 60 conflicting groups involving 143 of its 11,545 records. In each group,
a deterministic prediction based only on the complete visible input must
disagree with at least the less frequent label count. Summing those counts
gives a lower bound of 61 test-label disagreements
(Appendix Table~\ref{tab:pgb-edits}). This is a consistency bound, not 61 adjudicated
label errors or a measured model result. Split isolation addresses
cross-partition overlap; it does not resolve a test set that assigns
different answers to the same visible decision problem. Lossy summaries
and label noise remain competing explanations, so no replacement target
follows from the conflict alone.

ShARC asks models to answer rule-based questions or request clarification using the supplied rule, scenario, and dialogue history. Grouping identical visible inputs finds one conflicting-target group in train and one in dev, involving five records in total. In the dev witness, one reference answers ``No'', while the other asks for clarification; only a hidden annotation distinguishes the records (Appendix~\ref{sec:sharc}). The conflict does not establish which target should replace the other. Exposing that annotation would change the input.

No such exact-input conflict is found among ShARC's 8,276 test records.

The question itself can also rule out a reference answer that appears
plausible when reading only the policy. Two released ConditionalQA records
share a policy and scenario but ask different questions
(Case~\ref{case:cqa-companions}). One question explicitly excludes two answer
categories and supplies a permission premise. Its reference nonetheless
retains the excluded categories and an already-satisfied condition. The
paired record shows when those answers and conditions are appropriate.

\begin{figure}[!t]
\begin{auditcase}[unbreakable]{ConditionalQA: applying the question's exclusions and premises}{case:cqa-companions}
\casesource{Released \mbox{\artifact{dev-284}} and \mbox{\artifact{dev-242}}, with the shared
policy \emph{Raise a grievance at work}~\cite{conditionalqa}. Scenario and answer sets condensed; questions and quoted policy text verbatim.}
\casefield{Shared scenario}{Lucy has raised a grievance about poor working
conditions, and her employer has arranged a grievance meeting.}
\casefield{Policy}{A colleague or trade union representative can accompany
the worker. The policy adds: ``You may also be able to bring a family member
or Citizens Advice Bureau worker, depending on the HR procedure where you
work.''}
\casefield{\texttt{dev-284} question}{``Other than a colleague and her trade
union representative, who else can accompany her to the meeting if our HR
policy allows them to?''}
\casefield{Stored answers}{Colleague, trade union representative, family
member, and Citizens Advice Bureau worker. Only the last carries the quoted
HR condition.}
\casefield{Matched control: \texttt{dev-242}}{The policy and scenario are
identical, but the question is ``Who can accompany her to the meeting?''
Its reference includes all four categories, with the HR condition on both
the family member and Citizens Advice Bureau worker.}
\caseaudit{}
\casefield{Supported local target}{For \artifact{dev-284}, retain only the
family member and Citizens Advice Bureau worker, both with empty condition
lists. The question excludes the other two categories and asks about these
companions under the premise that HR permits them. No additional condition remains beyond that premise.}
\casefield{Control judgment}{For \artifact{dev-242}, all four categories
are warranted. Because its question supplies no HR permission, the
conditions on the family member and Citizens Advice Bureau worker remain
necessary.}
\end{auditcase}
\end{figure}

Other ConditionalQA source readings have different outcomes. The review
finds an unconditional answer not supported by its document without establishing a
unique replacement, and rejects a proposed label flip as a false alarm
(Appendix~\ref{sec:cqa-semantics}). In RuleArena NBA,
one reference names a team absent from the case while the violation verdict
remains supported (Appendix~\ref{sec:nba}).

\section{Effects on scores and model responses}
\label{sec:repairs}

We examine two consequences of the audit findings. \emph{Fixed-output rescoring} keeps model inputs and recorded responses unchanged while correcting one evaluation component. Reference corrections preserve the parser and acceptance rule; evaluator corrections preserve the references. \emph{Changing supplied intermediate answers} tests whether the amounts provided as solved steps in OpenExempt affect newly generated responses. The policy, question, and other case facts remain fixed across conditions. Experimental details and additional results are provided in Appendix~\ref{sec:additional-studies}.

We use Haiku 5.5, GPT6 Luna, and Gemma 4 31B. We score one response per input, model, and condition, generated without access to the reference answer for the question being scored. For Tax and SARA, responses are generated using the input templates supplied by the benchmark authors. ConditionalQA and the OpenExempt experiment use our \textbf{Single-turn} configuration, a single tool-free call that requests a JSON answer. We compare effects only within a configuration.

\subsection{Rescoring unchanged responses}
\label{sec:tax-impact}

\begin{table*}[t]
\centering\small
\setlength{\tabcolsep}{3.5pt}
\begin{tabular}{@{}lrrrrrr@{\hspace{1.4em}}rrrrr@{}}
\toprule
 & \multicolumn{6}{c}{RuleArena Tax (accepted / 300)} & \multicolumn{5}{c}{LegalBench SARA numeric (accepted / 96)}\\
\cmidrule(r{1.4em}){2-7}\cmidrule(l){8-12}
Model & Original & Revised & Gain & Loss & Floor & Zero only & Original & Parse & Parse\,+\,zero & Gain & Loss\\
\midrule
Haiku 5.5 & 209 & 224 & 16 & 1 & 16 & 0 & 8 & 87 & 91 & 83 & 0\\
GPT6 Luna & 241 & 261 & 20 & 0 & 10 & 10 & 9 & 83 & 87 & 78 & 0\\
Gemma 4 31B & 195 & 220 & 25 & 0 & 22 & 3 & 8 & 84 & 88 & 80 & 0\\
\bottomrule
\end{tabular}
\caption{Fixed-output rescoring with unchanged inputs and responses.
\emph{Tax:} Revised applies only the line-22 zero floor and recomputes
downstream amounts; native numerical closeness is fixed. Floor and Zero only
partition the Tax gains: responses explicitly state the floor or report
a zero subtotal without stating the operation. \emph{SARA:}
References and the approximately 10\% nonzero-reference tolerance stay fixed.
Parse preserves signed decimals under the original acceptance formula;
Parse\,+\,zero adds exact-zero handling for zero references. Gain/Loss count
newly accepted/rejected responses against Original: Revised for Tax,
Parse\,+\,zero for SARA.}
\label{tab:tax-impact}\label{tab:sara-impact}
\end{table*}

The Tax zero-floor correction raises reference agreement for all three models (Table~\ref{tab:tax-impact}, left). All changed judgments occur among the 27 revised references. All 300 cases remain in each denominator. The Haiku--Gemma gap falls from 14 to four accepted cases; model order is unchanged.

The visible responses help interpret the gains: 48 of the 61 newly accepted responses explicitly apply the zero floor, while 13 reach a zero subtotal without stating the operation (Table~\ref{tab:tax-impact}). Haiku's sole newly rejected response already has a zero subtotal, but compensating calculations make its final answer match the original reference. Conversely, agreement under this scoped correction does not certify the full return, because a floored zero can mask other intermediate errors. The reference correction also leaves the native parser and other source assumptions unchanged (Appendix~\ref{sec:tax-run-details}).

Correcting the LegalBench SARA scorer raises accepted answers from 8--9 to 87--91 out of 96, with responses and reference answers unchanged (Table~\ref{tab:sara-impact}, right). Most of the gain comes from preserving decimal amounts during parsing. The exact-zero correction then adds four accepted answers per model on the zero-reference cases, leaves all 92 nonzero-reference judgments unchanged, and introduces no losses. Together, these corrections change the observed ordering from Luna first, with Haiku and Gemma tied, to Haiku followed by Gemma and Luna. These orderings describe the retained responses, not a stable ranking of model ability.

Keeping the decimal point while retaining the original formula
still rejects an exact zero for a zero reference and can accept 0.10
instead. Separating the two changes shows which gains come from extracting
the stated amount and which require fixing zero-reference acceptance.
The final scorer preserves the original tolerance on nonzero references.

In ConditionalQA, the official evaluator excludes questions with missing
predictions from the aggregate score. Our JSON interface produces such
missing entries: all models attempt 285 questions, but Haiku has no missing
IDs, Luna has 24, and Gemma has 53. A submitted empty answer
list remains a valid response form for an unanswerable question; it is not
a missing prediction. Rescoring the same files with missing IDs counted as
zero lowers Luna's and Gemma's scores (Table~\ref{tab:cqa-main-impact}).
Submitted answers retain their native per-question scores, and all
reference labels remain unchanged.

Answer EM averages the native per-question metric, which can give partial
credit across multiple reference answers. It should not be converted into
an integer count of correctly answered questions. The lower aggregates
reflect the restored population, not deteriorated responses. Scores for
Haiku's complete submission and all eight complete Code Prompting files
are unchanged (Appendix~\ref{sec:cqa-study}).

\begin{table}[!tbp]
\centering\small
\setlength{\tabcolsep}{4pt}
\begin{tabular}{@{}lrrrr@{}}
\toprule
 & \multicolumn{2}{c}{Scored questions} & \multicolumn{2}{c}{Answer EM (\%)}\\
\cmidrule(lr){2-3}\cmidrule(l){4-5}
Model & Original & Revised & Original & Revised\\
\midrule
Haiku 5.5 & 285 & 285 & 58.63 & 58.63\\
GPT6 Luna & 261 & 285 & 52.34 & 47.93\\
Gemma 4 31B & 232 & 285 & 58.02 & 47.23\\
\bottomrule
\end{tabular}
\caption{ConditionalQA fixed-output rescoring. All models attempt 285
questions. The revision counts missing prediction IDs as zero; submitted
answers retain their per-question scores. Answer EM is their native
per-question mean over the stated population. References remain unchanged.}
\label{tab:cqa-main-impact}
\end{table}

\subsection{Effects of supplied intermediate answers}
\label{sec:oe-impact}

We use all 69 records containing federal solved-step values from the 100-record stage-4 optimal-exemption (OE4) component. We replace the original choice of federal or state exemptions and allocation with a question asking for the smallest total asset value that must remain unprotected from creditors after applying federal exemptions (the minimum non-exempt value). For this experiment, we specify that the spouses file jointly and both use federal exemptions. Each owns half of every listed asset, with no debts secured against those assets. The list includes all their assets, and the applicable exemptions are specified. The targets and step revisions follow these added premises; the experiment evaluates this amended task.

For each condition, we generate a new response with the policy, question,
and other case facts fixed: \emph{Released} supplies the released
intermediate amounts, \emph{Revised} supplies maxima recomputed for each
asset and exemption under the stated assumptions, and \emph{None} supplies
no intermediate amounts. A pair's maximum assumes that other pairs have
not yet consumed the shared allowances; the maxima cannot generally all be
used at once. The final target therefore comes from a joint allocation
subject to those shared limits. Before generation, separate allocation and
decimal calculations agreed on all 69 targets
(Appendix~\ref{sec:oe-step-design}).

\begin{table}[!tbp]
\centering\small
\setlength{\tabcolsep}{4pt}
\begin{tabular}{@{}lrrrrr@{}}
\toprule
 & \multicolumn{3}{c}{Accepted / 69} & \multicolumn{2}{c}{Changes}\\
\cmidrule(lr){2-4}\cmidrule(l){5-6}
Model & Released & Revised & None & Gain & Loss\\
\midrule
Haiku 5.5 & 50 (3) & 68 (18) & 69 (19) & 18 & 0\\
GPT6 Luna & 66 (16) & 68 (18) & 69 (19) & 3 & 1\\
Gemma 4 31B & 65 (16) & 64 (15) & 69 (19) & 2 & 3\\
\bottomrule
\end{tabular}

\caption{OpenExempt: one new response per case and condition, with released, revised, or no intermediate amounts. Acceptance requires the specified JSON format and an amount within \$0.005 of the target. Parenthesized counts cover the 19 cases where treating released amounts as caps changes the minimum non-exempt value and are included in the totals. Gain/Loss are newly accepted/rejected cases, Released to Revised.}
\label{tab:oe-intervention}
\end{table}

Revising the supplied steps changes accepted answers from 50 to 68 for Haiku, 66 to 68 for Luna, and 65 to 64 for Gemma (out of 69; Table~\ref{tab:oe-intervention}). With no supplied amounts, all three models achieve 69 of 69 accepted answers in these single generations. Luna's sole new rejection gives the correct amount in prose but fails our JSON format. Gemma's three new rejections satisfy the format but give incorrect amounts.

The targets comprise 20 positive residual values and 49 zeros; the None
condition is fully accepted in both groups (Appendix Table~\ref{tab:oe-step-controls}).
Thus its aggregate success is not confined to cases with no value left
unprotected.

Before generation, we identified 19 cases in which treating the released amounts as upper limits changes the minimum non-exempt value. This subset contains 16 of Haiku's 19 rejected answers under the released-step condition. Released-step outputs match the optimum under those ceilings in five Haiku, three Luna, and zero Gemma cases (Appendix Table~\ref{tab:oe-step-controls}); this diagnostic does not explain every rejection. Case~\ref{case:oe-step-response} in the appendix shows a response using a released amount in its final calculation. The four cases with unchanged supplied steps remain correct across all conditions, helping distinguish the changed intermediate information from the shared policy and case.

Native diagnostic cases overlap this experimental population, but the
added premises determine the revised amounts and targets. Individual
experimental revisions or response changes cannot all be attributed to
adjudicated native defects.

\section{Benchmark revisions}
\label{sec:implications}

\textbf{Revisions that permit rescoring.} Different benchmark defects require different forms of reevaluation. If the policy, question, case, and decision problem remain unchanged, a supported reference or evaluator correction allows the same responses to be rescored under both versions. The Tax zero floor and SARA numeric parsing and exact-zero rule correct identified defects. ConditionalQA applies a stated full-coverage rule: missing submissions receive zero. Its revised scores therefore reflect all reference questions rather than only those with submitted predictions. BAR provides all 300 Tax reference pairs with unchanged native inputs, alongside the SARA numeric and ConditionalQA scorers, which retain their original labels.

\textbf{Revisions that require new responses.} Restoring omitted source text or changing the decision problem requires new predictions because the input changes. Restoring a missing rule gives the model information needed to answer the original question from the supplied policy (Appendix Case~\ref{case:lb-context}). For RuleArena Airline, adding a minimum-cost objective to resolve the baggage-allocation ambiguity specifies which allocation is preferred \citep{nair}, defining an amended task rather than merely correcting a label. These revisions support different evaluations: one tests decisions with more complete policy information, while the other tests decisions under a newly specified objective.
The historical controls in Appendix~\ref{sec:historical-controls} distinguish
policy amendments from upstream repairs that preserve the decision problem.

\textbf{Revisions that need further review.} Other targets remain conditional or unresolved. Correcting OpenExempt's valuation routine yields candidate references conditional on the authors' annotations of which exemptions apply; those annotations still need semantic review. Records without a supported target remain unresolved; excluding them changes the evaluated population. Workflow findings and unresolved conflict groups remain audit evidence. Across these categories, BAR records revisions separately by benchmark component and source version, with evidence and use guidance. It distinguishes supported corrections, conditional candidates, unresolved findings, and upstream repairs. Any exclusion-based reevaluation must report original and retained denominators.

\section{Related work}

Prior audits identify label errors affecting model selection \citep{northcutt2021} and document MMLU errors through manual review and reevaluation \citep{gemammlu2025}. HLE-Verified separates verified, revised, and uncertain items \citep{hleverified2026}; a text-to-SQL audit shows that correcting questions and reference queries can change method rankings \citep{textsql2026}. Studies of task selection, aggregation, and construct validity examine what benchmark scores measure \citep{dehghani2021,bean2025}.

BenchGuard audits instructions, environments, reference programs, and evaluation logic in agent benchmarks \citep{benchguard2026}. UTBoost augments tests and repairs the parsing of test logs to expose incorrect patches accepted by SWE-Bench \citep{utboost2025}. ELT-Bench-Verified refines evaluation scripts and excludes columns with unreliable reference values before reassessing data engineering agents \citep{eltverified2026}.

For policy-based tasks, \citet{verma} identify spurious patterns and follow-up reference limitations in ShARC; our exact-input checks test target consistency. \citet{nair} clarify RuleArena Airline's allocation ambiguity. The previously reported LegalBench parsing and TheAgentCompany budget issues are attributed in Appendix Cases~\ref{case:lb-scorer} and~\ref{case:tac-budget}; our rescoring and runtime checks establish their effects under the stated conditions. We trace errors shared by reference generation and replay, inherited through supplied steps, or concealed by accepted program outputs. Our experiments separate effects on scoring from effects on subsequent decisions.

\section{Conclusion}

We audit nine public policy-based decision benchmarks and find that responses supported by the supplied policy and case can be rejected, while incomplete decisions can be accepted. Correcting references and evaluators changes measured agreement and observed model differences even when responses remain fixed. In our amended task, faulty intermediate answers can affect later decisions, and revising them affects models differently. Reference replay shows that agreement among benchmark components is insufficient when they reproduce the same error. Therefore, interpreting scores as policy-reasoning performance requires that model-visible inputs, supplied steps, references, and scoring reflect the same policy, question, case, and decision requirements. BAR preserves audit evidence and versioned revisions for reevaluation.

\section*{Limitations}

Audit coverage varies by benchmark and version. Some checks cover every record in a release; semantic reviews cover selected cases and use model assistance without domain experts. We may miss other error types or affected records, so unflagged records are not necessarily correct. Selected cases cannot establish an error rate for an entire benchmark. We identify specific failure mechanisms but do not systematically study how benchmark development and review processes produce or overlook these defects.

Our experiments use three models without tools and one retained response per input and condition. Score changes and model differences may vary with other models, prompts, or repeated generations. The OpenExempt experiment specifies ownership and eligibility and uses only federal exemptions; it does not fully repair the native task. WorkArena++ and TheAgentCompany checks use constructed inputs or states and do not measure failure frequency in live agent runs. TheAgentCompany's budget reference prices remain unverified.

We have not systematically re-evaluated prior studies using these benchmarks or measured effects on deployed systems. Saved responses can be rescored after reference or scorer corrections; changed inputs require new responses. Missing outputs or experimental records may prevent reassessment. BAR addresses known defects, but some corrections remain conditional on unresolved eligibility or case facts. Revised assets may still contain other errors.
\phantomsection
\addcontentsline{toc}{section}{References}

\bibliography{bibliography-control,references-display}

\clearpage
\raggedbottom
\appendix
\clubpenalty=10000
\widowpenalty=10000
\tcbset{lines before break=8}
\section*{Appendix}
\addcontentsline{toc}{section}{Appendix}
\etocdepthtag.toc{appendix}
\begingroup
\small
\etocsettagdepth{main}{none}
\etocsettagdepth{appendix}{subsubsection}
\etocsettocstyle{\subsection*{Contents}}{}
\tableofcontents
\endgroup
\medskip
\section{Audit scope and methodology}
\label{sec:versions}

\subsection{Task contracts, selection, and source families}
\label{sec:task-contracts}

The source families discussed here differ in the decisions they ask a model
to make and the information they expose (Table~\ref{tab:tasks}). Judging a
supplied trajectory is a static task, distinct from executing its actions.
In OpenExempt, valuation reports an independent maximum for each
asset--provision pair. Later decomposition tasks supply those values as
solved steps. Optimal-exemption stage 4 (OE4) asks for a federal or state
exemption choice and an estate-wide allocation, where pairs compete for
shared allowances. The native consistency check compares a pair's supplied
maximum with its allocation in that later reference.

\begingroup
\tablefont
\begin{table*}[!t]
\centering\small
\begin{tabular}{@{}L{.17\textwidth}L{.39\textwidth}L{.38\textwidth}@{}}
\toprule
Component & What the model receives and produces & How the target is scored\\
\midrule
RuleArena Tax / Airline & Supplied tax instructions or baggage rules and a
case; output a signed tax amount or the total travel
cost, including ticket and baggage fees. & Extracted numerical answer
compared with the reference using numerical closeness.\\
\rowcolor{black!6}
RuleArena NBA & League rules and a transaction scenario; decide whether a
violation occurs and identify its operation and team. & Decision and location
matching against the stored operation and team.\\
OpenExempt Exemption Valuation (EV) & Supplied exemption statutes and a debtor case; identify
applicable clauses and the maximum protectable value for each asset--clause
pair independently. & Asset-level F1 and numerical error; the paper describes
a 5\% amount tolerance and fuzzy asset-name alignment.\\
\rowcolor{black!6}
SARA in DeonticBench & Edited tax statutes and case facts; output a tax amount
or the truth of a statutory proposition. A Prolog reference program accompanies
each record. & Numerical or boolean answer agreement under the authors'
documented output-scoring rule, which tolerates rounding for amounts and is
also used after model inference.\\
USCIS in DeonticBench & Policy text and facts from immigration appeals; output a
disposition such as Accepted or Dismissed. & Disposition agreement; reference
program execution is an additional artifact to inspect.\\
\rowcolor{black!6}
ShARC & A rule passage, user question, scenario, and dialogue history; output
Yes, No, Irrelevant, or a follow-up question. & Four-way decision accuracy
(follow-up is the More class) and BLEU for generated follow-up questions.\\
ConditionalQA & A policy document and scenario-bearing question; output answer
text and any conditions on its applicability, or an empty answer list. &
Answer EM/F1 and condition-aware scores; boolean and extractive categories
are also reported.\\
\rowcolor{black!6}
LegalBench SARA &
A statute snippet or bundle, case narrative, and entailment or amount question. &
Balanced exact-match classification or the numeric acceptance function.\\
PolicyGuardBench &
A supplied policy and fixed trajectory summary; output violation/no violation. &
Binary accuracy and F1 are reported; exact published harness binding unresolved.\\
\rowcolor{black!6}
Selected WorkArena++ / TheAgentCompany tasks &
Accessible procedures, a goal, and service state or dialogue; perform and report a decision. &
Task-specific state/message checkpoints; the inspected predicates concern
scheduling gaps and budget decisions.\\
\bottomrule
\end{tabular}
\caption{Tasks and scoring contracts. These describe the released tasks; the
smaller populations and checks performed here appear in Table~\ref{tab:coverage}.
EM denotes exact match and F1 denotes token-overlap F1 where applicable.}
\label{tab:tasks}

\end{table*}
\endgroup

\begin{table*}[!t]
\small
\tablefont
\begin{tabularx}{\textwidth}{@{}L{.17\textwidth}Y Y@{}}
\toprule
Family & Actual evaluation use & What the use establishes\\
\midrule
RuleArena & Policy-as-Logic evaluates all three domains~\cite{nair}. &
Task use; our Tax study separately holds its own input and scoring interfaces constant.\\
ConditionalQA / ShARC & Code Prompting reports ConditionalQA results;
its experiment repository includes ShARC runs~\cite{codeprompt}. &
Named task use, with different answer and clarification contracts.\\
LegalBench & Chain of Logic evaluates diversity, personal jurisdiction,
and J.Crew tasks~\cite{chainlogic}. &
Task-specific adoption, not verified use of the SARA scorer audited here.\\
PolicyGuardBench & UniGuard reports external benchmark
evaluation~\cite{uniguard}. &
Independent use; the evaluated original/strict revision is unresolved.\\
WorkArena++ & TeleTune evaluates a selected L2 cohort~\cite{teletune}. &
Adapted native tasks; exact audited-class membership and revision unresolved.\\
TheAgentCompany & Is Bash All You Need? evaluates workplace
agents~\cite{bashinterfaces}. &
Independent adapted use; exact selected tasks and evaluator versions unresolved.\\
\bottomrule
\end{tabularx}
\caption{Examples of verified reuse motivating coverage.}
\label{tab:adoption}

\end{table*}

\paragraph{Selection and verified reuse.}
\label{sec:why}

Several of these artifacts already enter later evaluations
(Table~\ref{tab:adoption}). A defect in a
reference or grader matters because later comparisons can inherit it.
DeonticBench's inclusion is motivated by its released reference programs and
available evidence. Original SARA supplies a lineage anchor.

OpenExempt's paper describes a generated legal-workflow benchmark with
inspectable policies, applicability annotations, and reference code~\cite{openexempt}.
These artifacts motivate our audit of its generated references; broad
independent reuse remains \mbox{unestablished}.

Outside the nine-benchmark audit, SOP-Bench supplies a boundary case for
single-reference evaluation when the policy permits investigator discretion.
Selected workflow validators extend
the audit to operational decisions.

\subsection{Inspected populations and verification coverage}
\label{sec:expanded-coverage}

The selection record covers 63 source/interface verification units, including
adaptations and boundary checks, not 63 independent datasets. All 162
LegalBench task entries received catalog screening; only selected components
received detailed checks. Table~\ref{tab:coverage} lists the populations
supporting the findings presented in this paper, rather than every screened
unit. Selection was purposive and favored accessible sources, limiting
generalization beyond these populations.

\begin{table*}[!t]
\centering\small
\begin{tabularx}{\textwidth}{@{}L{.19\textwidth}L{.23\textwidth}Y@{}}
\toprule
Component & Inspected population & Check and coverage boundary\\
\midrule
RuleArena Tax & All 300 cases &
Reference replay; three-model official-interface rescoring and trace review
of all 62 changed judgments; source-dependency checks.\\
\rowcolor{black!6}
RuleArena NBA & 216 annotated cases; selected localization witness &
Policy, substring-grader and rule-attribution vocabulary checks;
no complete decision-solver replay.\\
RuleArena Airline & 300-case release; allocation convention &
Supplied policy, task and reference-code inspection; prior-reported
objective ambiguity, not a count of wrong labels.\\
\rowcolor{black!6}
OpenExempt EV1 & 100 test records each in Basic, Intermediate, Advanced &
Replay and conditional amount/allocator checks; selected semantic witnesses,
not complete semantic coverage.\\
OpenExempt decomposition & 100 test records each in non-exempt-asset stage 4 (NA4) and optimal-exemption stages 4 and 5 (OE4, OE5) &
Intermediate-input inspection, source replay, and bounded consistency checks;
same source family as EV1.\\
\rowcolor{black!6}
DeonticBench & 100 SARA numeric, 276 SARA binary, 242 USCIS records &
Screening of all records; source and execution checks of selected allegations.
Airline and Housing excluded.\\
LegalBench SARA & 100 numeric; 276 entailment rows &
Full structural mapping and numeric checks; selected complete-input semantic
witnesses. Performance rescoring uses 96 distinct numeric test cases.\\
\rowcolor{black!6}
ShARC & 21,890 train; 2,270 dev; 8,276 test &
Exhaustive within-split grouping of exact model-visible inputs.\\
ConditionalQA v1.0 & 2,338 train; 285 dev; 804 public test; 652 documents &
Structural checks and evaluator replay; targeted semantic review of 41
distinct train/dev cases. Test labels unavailable.\\
\rowcolor{black!6}
PolicyGuardBench & 59,997 rows per original/strict release &
Complete-input grouping and split comparison.
Both releases contain the same labeled multiset, not independent corpora.\\
WorkArena++ & One scheduling class, L2/L3 views &
Constructed-state native-method witnesses; no suite-wide claim.\\
\rowcolor{black!6}
TheAgentCompany & Budget, single-password, and batch-password tasks &
Budget omission verified through the official scoring chain on constructed
replies and trajectory; HR user-coverage omission checked statically.
Budget prices unverified; no measured agent-frequency effect.\\
\bottomrule
\end{tabularx}
\caption{Inspected populations and checks. Counts use each row's stated unit
and may share provenance. Workflow findings concern constructed states and
grader inputs, not live agent episodes.}
\label{tab:coverage}\label{tab:expansion}
\end{table*}

The selection record also retains checks without a confirmed defect and
unresolved candidates. Reconstructing the stated rules for LegalBench's
diversity tasks finds no label mismatch under the documented convention;
selected hearsay cases lack a settled semantic judgment. Discount references
in selected AppWorld tool-use tasks agree with the frozen initial states,
but do not establish policy-sensitive outcomes under the inspected exposure.
The selected COLIEE legal-entailment release and clinical-reportability
candidates retain source-access gaps. Original SARA v3 program and annotation observations do not establish a released-label error rate. ToolEmu's hidden expectations belong to its declared threat model and input conditions. These outcomes
limit the audit's claims; they are not additional defective \mbox{benchmarks}.

\subsection{Source review, dispositions, and independent readings}
\label{sec:screening}

Source review compares candidate findings with frozen policies, case fields,
reference programs, and evaluators. These comparisons were model-assisted,
without separate human or domain-expert adjudication. Some used Codex
without recording its underlying model identifier. Tax trace coding used
one pass with scores visible and no independent recoding.
Reviewers of candidate findings and proposed repairs saw those findings
alongside the source material. Repair reviews by Sol 6.1 and Claude Opus 5.5
were independent between readers, but not blind to the candidates.
These reviews do not constitute blind, independent adjudication of all
63 verification units.

GPT6 Luna screened all 618 selected DeonticBench records; some screening
records contain nonliteral quotations.
The paper uses only source-backed observations and retained execution evidence.
The candidate-to-disposition mapping is incomplete, so no rejection rate
can be inferred.

\paragraph{ConditionalQA source review.}
\label{sec:model-readings}\label{sec:reading-followup}
We selected 16 train and eight dev cases at equally spaced positions in
released order among answerable questions with one yes/no answer and no
annotated conditions. Source-linked candidates and controls add 17 distinct
cases. These selections support targeted findings, not an estimate of
semantic error prevalence. Adjudication compares the reference answers
and their conditions with the supplied policy, question, and scenario,
considering missing facts and competing interpretations.

The first review read complete documents for eight of the 24
cases and the question, scenario, reference, and cited passages for the
other sixteen. Separate readings by GPT6 Luna, GPT-6 Astra, and Claude
Opus 5.5 received the complete policy, question, and scenario for those
same 24 cases without labels, prior judgments, or other readers' answers. Disagreements
about permission, procedure, and missing facts informed the case review;
model agreement did not establish a correct target.

\FloatBarrier
\section{Detailed findings and revisions}
\label{sec:more-cases}
\label{sec:additional-targets}
These cases separate source excerpts from audit calculations and judgments.
Quotation marks indicate exact source wording; condensed source fields are
labeled. Input-support audits compare each target with the policy and facts
visible to the model, distinguishing missing evidence, conflicting targets,
and unsupported components of an otherwise unchanged decision.

\subsection{RuleArena}

\subsubsection{Tax: reference trace and input dependencies}
\label{sec:tax}

RuleArena Tax asks for a signed tax amount from supplied forms and case facts.
Its released reference calculates a tax subtotal without the zero floor required
by the form instructions. Replaying the pinned reference with only the floor
changed and all other inputs and defaults retained changes 27 of the 300 signed
amounts: three in complexity level 0, 11 in level 1, and 13 in level 2.
The original \texttt{numpy.isclose} answer check rejects all 27 revised reference
amounts. The remaining 273 amounts are unchanged. Case~\ref{case:tax-floor}
gives the general calculation and all-case control; Case~\ref{fig:opening-witness}
shows the concrete example.

\begin{auditcase}{RuleArena Tax: checking the reference revision}{case:tax-floor}
\casesource{Frozen RuleArena form instructions and reference code~\cite{rulearena}
(\href{https://github.com/SkyRiver-2000/RuleArena/tree/59f60f6b4eb93a9b6d187ec781349d4f15618903}{official repository}).}
\casefield{Rule and implementation}{After subtracting credits, the form
requires a negative subtotal to be entered as zero. The final-answer program
omits this step; the rule-level analyzer includes it.}
\caseaudit{General reference relation and all-case controls.}
\casefield{Calculation}{Let \(t\) be the tax subtotal before credits on
Form 1040 line 18, \(c\) the credits on line 21, \(o\) other taxes, and \(p\)
payments. The original and revised amounts are
\[
 \begin{aligned}
 a_{\mathrm{old}}&=(t-c)+o-p,\\
 a_{\mathrm{floor}}&=\max(0,t-c)+o-p.
 \end{aligned}
\]
Positive amounts are owed; negative ones are refunds.}
\casefield{All-case control}{For all 300 cases, the two programs use the same
inputs. Their outputs satisfy
\(a_{\mathrm{floor}}=a_{\mathrm{old}}-\min(0,t-c)\): amounts are unchanged
when \(t-c\geq0\), and only the negative subtotal is removed otherwise.
Changed reference fields are restricted to line 22 and its downstream totals.}
\casefield{Revision boundary}{All other reference logic and defaults,
including the zero default for an unavailable Schedule 3 input, are retained.}
\end{auditcase}

The reference uses a low self-employment-earnings threshold that the supplied
packet does not fully expose and substitutes zero for an unavailable Schedule 3
input. All 300 reference calculations use the Schedule 3 default, which the floor
repair retains. The low-earnings branch is reached in 21 cases: five changed and
16 unchanged by the floor repair. These counts describe assumption use and
branch execution. The repair records flag both dependencies, including those
in the 273 floor-unchanged cases.

\begin{auditcase}{Tax: an input dependency beyond the zero floor}{case:tax-residual}
\casesource{Frozen task \artifact{tax/comp_2/74}, supplied Schedule SE,
and reference program.}
\casefield{Policy and case}{This step computes the self-employment tax used
in the final tax calculation. The case reports a business loss of \$1,706
on Schedule C.
Schedule SE line~4c says to retain line~3 when it is not positive.
Lines~10 and~11 refer to ``line 4'', although the supplied form shows
line~4c. They apply 12.4\% and 2.9\%, and line~12 adds them.}
\casefield{Reference branch}{For a base below \$400, the program sets
self-employment tax and its deductible half to zero. The supplied text
does not include this branch.}
\caseaudit{Scope of the reference revision.}
\casefield{Consequence}{Reading ``line 4'' as line~4c gives
\(-1{,}706(0.124+0.029)=-261.018\), while the reference returns zero.
This is a conditional intermediate amount. The line~22 repair retains
this separate dependency; remaining disagreements cannot all be attributed
to model error.}
\end{auditcase}

\subsubsection{RuleArena NBA: localization and rule attribution}
\label{sec:nba}

RuleArena NBA asks whether a proposed transaction violates league rules
and, if so, which operation and team are responsible.
In the inspected witness, the stored team does not appear in the case;
the team supported by the case and policy fails the frozen substring grader,
while the violation verdict stands (Case~\ref{case:nba-team}). An apparent
clash between percentage provisions is resolved by the policy's explicit
sign-and-trade exception. The intended single-transaction treatment is also
documented in \href{https://github.com/SkyRiver-2000/RuleArena/issues/3}
{RuleArena issue~\#3} and its reply.

\begin{auditcase}{RuleArena NBA: a team absent from the case}{case:nba-team}
\casesource{Released RuleArena case, reference, policy, and
grader~\cite{rulearena}; condensed.}
\casefield{Record}{\artifact{comp_1}, index 62.}
\casefield{Operation B}{Team Q signs Player G with 6\% annual increases and
immediately trades him to Team F.}
\casefield{Supplied policy}{Article VII, Section 5(a)(2) expressly excludes
sign-and-trade contracts under Section 8(e)(1) from its 8\% allowance and
directs them to Section 5(a)(1), which limits annual increases to 5\% of
first-year salary.}
\casefield{Stored team}{H.}
\casefield{Grader (condensed)}{The frozen grader searches the response for the expected operation and team
text.}
\casefield{Neighboring records}{Indices 60 and 61 use Team H in the same template.}
\caseaudit{}
\casefield{Applicability}{Operation B proposes a sign-and-trade. Its
6\% annual increase therefore exceeds the applicable 5\% limit.}
\casefield{Localization}{Q offers the contract that exceeds this limit.
No Team H appears in the case. The grader accepts an answer naming H
and rejects the otherwise identical answer naming Q. The neighboring records
are consistent with an incomplete rename.}
\casefield{Adjudication}{The stored team is unsupported, while the
salary-increase rule supports identifying Q in operation B.
The violation verdict remains unchanged. Whether the receiving team F
also violates the separate Room (available salary-cap capacity) requirement
remains unresolved.}
\end{auditcase}

The separate rule-extraction evaluator uses 55 Boolean
fields. Nine rule identifiers in the annotations are absent from that
vocabulary, accounting for 54 distinct case--identifier occurrences across
46 of 216 cases. Literal identifier matching counts those occurrences as
missed rules even if every expressible annotated rule is extracted.
The per-rule report, in turn, lists only identifiers in the schema.
This is a rule-attribution scoring mismatch, not 46 wrong final verdicts;
we do not measure its effect by rescoring model outputs.
Some identifiers resemble aliases, but a repair must verify their meaning
rather than merge them by spelling alone.

\subsubsection{RuleArena Airline: an unstated allocation objective}
\label{sec:airline}

Airline asks for total travel cost, including ticket and baggage fees.
The supplied rules leave unclear which listed bags receive complimentary
allowances and occupy numbered fee positions.
Case~\ref{case:airline-allocation} shows how assigning the same complimentary
baggage allowances to different bags changes the total charge.
Nair et al.\ previously identified this
problem and added a minimum-cost directive to their ASP formalization
\citep[Section~4.2]{nair}. RuleArena
\href{https://github.com/SkyRiver-2000/RuleArena/issues/5}{issue~\#5}
separately reports the ordering concern.

\begin{auditcase}{RuleArena Airline: the reference chooses an allocation}{case:airline-allocation}
\casesource{Frozen RuleArena baggage policy, task prompt, and reference
calculation~\cite{rulearena}; \mbox{condensed}.}
\casefield{Supplied task}{Compute total travel cost under the baggage
rules. Complimentary status can affect a bag's weight allowance; base fees depend on bag position.}
\casefield{Reference convention}{The program computes each bag's surcharge
savings from complimentary status, orders bags by decreasing savings,
and then assigns numbered fees. The supplied task does not explicitly
state this allocation convention.}
\caseaudit{Trace the reference's objective.}
\casefield{Allocation witness}{In the inspected domestic First-class case
(\artifact{comp_0}, index 2), four of the five listed items are checked.
Using the original item numbers and the frozen calculator's baggage
assumptions, assigning complimentary status to items 2 and 3 gives a total
cost of \$1,019; assigning it to items 2 and 5 gives \$949.
The reference calculation selects the latter. Both totals include the same
ticket price and total base baggage fees.}
\casefield{Why the totals differ}{Item 5 weighs 69 lb. Complimentary status
raises its weight allowance from 50 to 70 lb, reducing its applicable
surcharge from \$100 to the \$30 oversize fee. Item 3's surcharge is
unchanged, so the reassignment saves \$70.}
\casefield{Consequence}{An answer based on another allocation need not
show a failure to apply the stated fees. An explicit objective is needed
to make the intended decision problem clear.}
\casefield{Scope}{The source and saved calculations establish the allocation
convention and this cost difference, not global optimality or an error rate
over the 300 cases. Adding an
objective changes the task and requires new predictions; it does not
retroactively validate old scores.}
\end{auditcase}

\subsection{OpenExempt: references and downstream inputs}
\label{sec:oe-recurrence}

OpenExempt's exemption-valuation task (EV1) asks which provisions apply to
each asset and the maximum value each provision can protect from creditors.
We check whether the reference computation produces that maximum
independently for each asset--provision pair, conditional on the source
applicability annotations, and whether later tasks receive consistent
amounts as solved steps.
The inspected paper versions do not identify the exact source revision used
for every reported model result.

\paragraph{The shared generation path.}
In the traced Arizona witness, the source annotation encodes the combined
married allowance for a vehicle and a claim count of two, consistent with the
supplied policy. The EV caller, however, invokes a one-claim allocation helper
once and publishes its result as the maximum, whereas the author's
estate-level search can claim the same allowance again. Where an asset exceeds
a single allowance and its annotation permits a second claim, the EV reference
therefore records one claim. Because generation and replay use the same code,
the released label reproduces exactly. Case~\ref{case:oe-arizona} traces this path. Table~\ref{tab:openexempt} counts where
the same program diagnostic applies across suites.

\begin{auditcase}{OpenExempt: an incomplete spousal maximum}{case:oe-arizona}
\casesource{Released Advanced EV1 record and frozen source~\cite{openexempt}
(\href{https://huggingface.co/datasets/SergioServantez/OpenExempt/tree/842cedec770ff2e45636fb5cb526ba0708ceb908}{released dataset};
\href{https://github.com/servantez/OpenExempt/tree/ec6fa2547431096588ce2d3ef107b8adc33cd4fe}{official code}); policy, case,
program and paper contract condensed.}
\casefield{Record}{{\raggedright\artifact{ac_exemption_valuation_1_}\allowbreak\mbox{\artifact{test_task_28}}.\par}}
\casefield{Task}{Report the maximum protectable value for each
asset--provision pair, considered independently.}
\casefield{Assets}{A \$48,900 Lincoln Navigator and a \$930 jacket.}
\casefield{Supplied policy (condensed)}{Arizona \S33-1121 permits spouses to
combine exemptions on the same property. Section 33-1125(8) provides \$15,000
for one vehicle; \S33-1125(1) provides \$500 for clothing.}
\casefield{Ownership instruction (condensed)}{For a married couple, assume joint
ownership with equal undivided interests, unless explicitly stated otherwise.}
\casefield{Hidden case metadata}{The debtors are represented as married.}
\casefield{Vehicle annotation}{Combined married allowance: \$30,000.
Item-claim count: two.}
\casefield{Released references}{Vehicle: \$15,000. Jacket: \$930.}
\caseaudit{}
\casefield{Input observations}{The visible prose does not explicitly state
that the debtors are married, and no separate lien or equity field is
provided. Equal joint ownership therefore follows from the conditional
ownership instruction together with the hidden married metadata.}
\casefield{Calculation}{Using the source's married metadata, conditional
equal-interest instruction, and represented asset values, the combined
vehicle maximum is \$30,000, whereas the released reference is \$15,000:
\[
\begin{aligned}
2\min(48900/2,15000)&=30000,\\
2\min(930/2,500)&=930.
\end{aligned}
\]}
\casefield{Clothing control}{One spouse's \$465 jacket interest is below the
\$500 allowance; the released \$930 therefore rules out a uniform single-person
convention.}
\casefield{Program trace}{The allocation routine returns one claim per
call. The exemption-valuation routine calls it once per asset--clause pair
and publishes \$15,000 as the maximum. The estate-level search can call
it again for the same provision; a second call returns the remaining
\$15,000.}
\casefield{Below-cap control}{The one-call helper fully protects the \$4,275
vehicle in Advanced task 17, leaving its reference unchanged.}
\casefield{Adjudication}{An implementation/reference inconsistency under the
source's documented task and represented facts. The defect lies in the EV
caller; the policy and the correctly encoded allowance are retained.}
\casefield{Remaining uncertainty}{The input limitations above require explicit
conventions in a revised task; they do not reconcile the source's own \$30,000
allowance with its one-call maximum. Whole-case adjudication remains open.}
\end{auditcase}

\paragraph{Recurrence across difficulty settings.}
We replay all 100 EV1 test records in each of the Basic, Intermediate,
and Advanced suites; every released answer is reproduced. To test the
one-call mechanism, we initialize the source allocator separately for
each annotated asset--clause pair with an item-claim count. We allocate
one claim and then request a second claim against the remaining asset
value. Table~\ref{tab:openexempt} counts positive second allocations.
The check uses the source's allowances, claim counts, and represented
debtor status; it does not independently establish statutory eligibility.

\begin{table}[!htbp]
\centering\small
\begin{tabularx}{\linewidth}{@{}Yrr@{}}
\toprule
Suite / represented status & Cases & \shortstack[r]{Positive second /\\item-count pairs}\\
\midrule
Basic / single & 100 & 0/24\\
Intermediate / single & 50 & 0/41\\
Intermediate / married & 50 & 14/24\\
Advanced / married & 100 & 84/124\\
\bottomrule
\end{tabularx}
\caption{OpenExempt EV1 program checks. The numerator counts pairs with a
positive second allocation, and the denominator counts all annotated pairs
with an item-claim count. These are observations of the source allocator,
not adjudicated legal-error rates. A zero numerator does not certify
eligibility or completeness of the released answer.}
\label{tab:openexempt}
\end{table}

\paragraph{A separate cap screen and its limits.}
A policy-based amount screen uses caps transcribed from the frozen policy,
while taking source-selected applicability as given. Its discrepancies are
conditional candidates, distinct from the program behavior above.
For example, Advanced EV1 task~9 values an Acura at \$8,200 and records
\$400 under Oregon \S18.345(1)(p). Combining two \$400 allowances on equal
half interests would give
\(\min(4{,}100,400)+\min(4{,}100,400)=800\).
However, \S18.345(3) requires joint judgment debtors who opt to combine
the relevant jointly owned property exemptions. Marital metadata does
not establish every prerequisite or the exercise of that option.
This screen therefore does not establish an unconditional \$800
replacement. Unflagged pairs and omitted provisions remain unverified.

\paragraph{Additional downstream checks and controls.}
We inspect three released Reasoning Decomposition files, each with 100 records:
non-exempt-asset calculation at step 4 (NA4), optimal schedules from step 4 to
step 5 (OE4), and schedules at step 5 (OE5).
Every NA4/OE4 prompt contains EV amounts presented as already solved
reasoning steps. Their 200 embedded answers exactly match native EV replay.
The source helper still has an additional positive claim in 92 pairs across
55 NA4 records and 98 pairs across 60 OE4 records. These program flags take source applicability as given.

For the 98 flagged OE4 pairs, we sum the terminal allocation entries for
the same asset and citation and compare that sum with the supplied EV
amount. In 38 pairs across 33 distinct records, the terminal allocation
is strictly larger (Case~\ref{case:oe-downstream}). These are a subset of
the 98 pairs in 60 records that permit a second claim: an available claim
need not be used in the terminal schedule. The comparison inherits the
source annotations and the EV per-pair-maximum contract; it does not
adjudicate each pair's legal prerequisites.

All 100 OE5 prompts supply a non-exempt total instead of an EV table;
the direct EV-input finding concerns NA4 and OE4. The three files contain
300 distinct structured cases after removing identifiers, rather than
matched variants of the same 100 cases.

\paragraph{Applicability remains a separate question.}
The preceding amount checks take the source's applicability annotations
as given. Case~\ref{case:oe-boundaries} examines whether the listed provisions
apply to the asset in the first place.
\begin{auditcase}{OpenExempt: business use and a personal-use restriction}{case:oe-boundaries}
\casesource{Released Advanced EV1 task 30; condensed policy, case, and label fields.}
\casefield{Case and reference}{The asset is a massage table explicitly needed
for a licensed practice. The released reference lists Illinois
\S12-1001(b) and \S12-1001(d) for this asset. The task assumes personal use
unless explicitly stated otherwise.}
\casefield{Supplied policy}{Clause (b) covers an equity interest in other
property; clause (d) covers implements, professional books, or tools of
the trade. A provision at the end of \S12-1001 restricts the section's
exemptions to personal rather than business use.}
\caseaudit{Compare the asset's stated use with the conditions on the listed
exemptions.}
\casefield{Adjudication}{On the ordinary reading, the explicit business
use conflicts with applying (b) under the section-wide personal-use
restriction. Clause (d)'s express coverage of trade tools creates a separate
interpretive tension, which remains unresolved. Correcting the amount
calculation cannot resolve whether a provision applies to the asset.}
\end{auditcase}

\subsection{DeonticBench: reference programs and output scoring}
\label{sec:deontic-detail}

Each selected DeonticBench record includes a Prolog program intended to
encode its policy and facts and derive answers from them. The artifact supports both
label-based answer scoring and reference-program execution or reuse. A
program defect can affect reuse even when the observed answer matches the label.

The selected components contain 100 SARA numeric, 276 SARA binary, and
242 USCIS records. We identify discrepancies in 20 distinct reference
programs: six SARA numeric, eight SARA binary, and six USCIS programs.
These counts describe affected programs, not independent defect types.

Seven programs have witnesses in the released cases themselves: three
produce inconsistent numerical or logical derivations, one mistranslates
a case fact, and three report errors when loaded. The remaining thirteen
contain policy-to-code discrepancies established through source comparison
or probes with modified inputs. Those observations do not establish
incorrect terminal answers for the released cases. No replacement
reference label is established by these checks.

Our replay first applies the authors' official entrypoint processing,
which supplies the query used to run each program and print an answer; this
processing introduces no new decision rules. It then compares the parsed
result with the label using the authors' documented output-scoring rule,
which is also used for executed model outputs.
For numerical cases this rule rounds the output and accepts an absolute
difference of at most one from the integer label, so the 618 accepted outputs
are not all exactly equal to their labels.
In the load-error cases, Prolog reports an error while loading the program,
but the entrypoint query still yields a parseable result that the scorer accepts.

Prior DeonticBench \href{https://github.com/guangyaodou/DeonticBench/issues/1}{issue~\#1}
and \href{https://github.com/guangyaodou/DeonticBench/issues/2}{issue~\#2}
report output--label mismatches. The latter describes a Harbor adapter that
truncates amounts and compares exactly. It would reject the reported
\texttt{tax\_case\_29} pair, 40740.32 versus 40741, which the authors'
rounding tolerance accepts. We have not executed that adapter; our counts
use the authors' rule, and the labels remain unchanged.

Case~\ref{case:deontic-tax} shows how compensating errors can conceal a
program discrepancy. Case~\ref{case:deontic-probes} shows unsupported additional derivations.

\paragraph{Qualification and execution scope for Case~\ref{case:deontic-tax}.}
In \artifact{tax_case_61}, Alice has gross income of \$102,268 in 2015
and takes the standard deduction. Bob has shared her principal place of
abode since 2012 and has no income. The supplied \S151(b) grants the taxpayer
an exemption, \S151(c) grants one for each dependent, and \S151(d)(1) fixes
that exemption at \$2,000. The supplied \S152(d) includes the father and
no-income conditions together with a qualifying-child exclusion.

Under the frozen represented facts, Bob meets the father and no-income
tests. Another taxpayer's deduction for Alice, a joint return by Bob, and
qualifying-child status for Bob are not established. The original SARA rules
treat such unproved exceptions as absent (negation as failure); the narrative
does not expressly negate them. The represented AGI is below every supplied
phaseout threshold, the lowest being \$150,000, and the 2018--2025
zero-exemption rule does not apply to 2015.

The original dependent query returns \texttt{[bob,bob]}. The father relation
and a separate qualifying-relative assertion each prove dependency for the
same person. Summing one allowance per proof produces the saved \$4,000
exemption and \$25,055.08 tax, accepted under the authors' scoring rule.
Removing only the duplicate relationship fact yields a \$2,000 exemption
and \$25,675.08 tax because Alice's allowance is still absent. This diagnostic
changes the program, not the released facts or label; no replacement label
is inferred from it.

\Needspace{18\baselineskip}
\begin{auditcase}{DeonticBench SARA: an accepted answer conceals extra tax derivations}{case:deontic-probes}
\casesource{Released reference program and label for
\texttt{s1\_a\_1\_ii\_pos} and its supplied statute;
facts and program clauses condensed.}
\casefield{Record}{\artifact{s1_a_1_ii_pos}.}
\casefield{Task and case}{Decide whether a stated tax amount follows from
the supplied facts and statute. Alice is married
under section 7703 for 2017 and files a joint return with her spouse; their
2017 taxable income is \$42,876. Proposition: ``Alice and her spouse have to
pay \$7208 in taxes for the year 2017 under section 1(a)(ii).'' Stored label:
\texttt{1}.}
\casefield{Supplied statute (condensed)}{Section 1(a)(ii) imposes \$5,535 plus
28\% of the excess over \$36,900 ``if the taxable income is over \$36,900 but
not over \$89,150''. Brackets (iii) and (iv) apply only to taxable income over
\$89,150 and over \$140,000, respectively.}
\casefield{Program (condensed)}{\texttt{taxpayer\_tax(Person, T)} derives
joint filing from the married and joint-return facts, reads taxable income,
applies a bracket clause, and rounds to whole dollars. The clauses for
brackets (i) and (v) test taxable income \(\le 36{,}900\) and
\(>250{,}000\); those for (ii)--(iv) test only the upper limits
\(\le 89{,}150\), \(\le 140{,}000\), and \(\le 250{,}000\). The verification
prints \texttt{Result: Entailment} if \texttt{taxpayer\_tax(alice, 7208)} has a
derivation, and \texttt{Result: Contradiction} otherwise.}
\caseaudit{Run the proposition check and enumerate the program's tax
derivations under the same facts.}
\casefield{Scored answer}{The original proposition check returns
\texttt{Entailment}, matching the stored label \texttt{1}. The amounts
below are separate tax derivations, not three accepted proposition answers.}
\casefield{Preserved probe}{With the facts unchanged,
\texttt{findall(T,taxpayer\_tax(alice,T),Ts)} returns
\texttt{[7208,5820,964]}.}
\casefield{Why the program still passes}{All three probe outputs answer the same
query: the rounded tax on Alice's joint return under the fixed facts. Only
bracket (ii) covers \$42,876, and it gives
\(5535+0.28(42876-36900)=7208.28\), rounded to 7208; this supports the
label. Because the program omits the lower limits, the bracket (iii) and (iv)
clauses also succeed, giving \(20165+0.31(42876-89150)=5820.06\) and
\(35928.5+0.36(42876-140000)=963.86\), rounded to 5820 and 964. The
original proposition check asks whether 7208 can be derived, not whether
it is the only amount the program can derive.
The authors' rule therefore accepts the output without testing whether
the program derives a unique supported tax amount.}
\casefield{Adjudication}{Program discrepancies with labels unchanged. The two
extra probe outputs are program derivations unsupported by the statute, not
replacement labels.}
\end{auditcase}

The nine supplied statutes are byte-identical to original SARA, and all
376 selected labels are unchanged in the adaptation. The findings concern
DeonticBench's reference programs.

\begin{auditcase}{DeonticBench USCIS: a delivery date encoded as a mailing date}{case:deontic-date}
\casesource{Released narrative and reference program for
\artifact{uscis_7c309e73}~\cite{deonticbench}; condensed.}
\casefield{Task}{Determine whether to accept or dismiss an
immigration-bond appeal.}
\casefield{Narrative}{The notice was mailed on November 9, 2023, requesting
delivery of the bonded individual on \mbox{December~11}.}
\casefield{Program}{The fact is
\texttt{form\_i340\_mailed\_on\_2023\_12\_11}.
The rule \texttt{bonded\_person\_absent\_on\_delivery\_date} uses this
fact together with the individual's entry into Canada and protected
status there. Its comment identifies December 11 as the delivery date.}
\caseaudit{Compare the roles assigned to the same dates.}
\casefield{Finding}{The fact's name and encoded date misstate the
narrative's mailing event. Its use in the delivery-date rule is consistent
with the requested delivery date, so this observation establishes a
mislabeled fact, not an incorrect date calculation or appeal disposition.
With the authors' entrypoint processing, the program returns Accepted,
matching the released label. No replacement disposition is established.}
\end{auditcase}

\subsection{LegalBench SARA}

\subsubsection{Numerical scoring diagnostics}
\label{sec:lb-scoring-diagnostics}

LegalBench SARA numeric asks how much tax is owed.
Case~\ref{case:lb-scorer} submits constructed answer strings to the
original scorer, testing whether amounts equal to the reference are
accepted and whether an amount 100 times smaller can also be accepted.
Decimal removal changes the amount being compared, and the
implemented inequality cannot accept a zero reference.

\begin{auditcase}{LegalBench SARA: exact amounts rejected by the numeric scorer}{case:lb-scorer}
\casesource{LegalBench~\cite{legalbench},
\href{https://github.com/HazyResearch/legalbench/blob/b46bf4ffae90524b2b72aaa30e7745fe9db64481/evaluation.py}{\texttt{evaluation.py}},
\texttt{evaluate\_sara\_numeric\_acc}; frozen train/test references in the
\href{https://huggingface.co/datasets/nguha/legalbench/tree/daec8237410aa23e3faf4bc41ad8b3a7e1696826}{released dataset}.}
\casefield{Declared task}{Report the amount first. The task asks how much
tax is owed, and the prompt does not forbid cents. The evaluator describes
acceptance within 10\% of the reference.}
\casefield{Implemented check}{Delete commas and decimal points, extract the
first digit run as a nonnegative integer \(\hat y\) (or zero if absent), and
accept against the numerical reference \(y^\star\) when
\[
\left|\frac{\hat y}{y^\star+0.1}-1\right|<0.1.
\]}
\caseaudit{Unchanged-source replay with constructed answer strings.}
\casefield{Representation witness}{For reference \texttt{\$259487}
(\artifact{tax_case_25}), the original scorer gives these results
for constructed answer strings:
\begin{center}
\begin{tabular}{@{}lrl@{}}
\toprule
Answer string & Parsed value & Accepted\\
\midrule
\texttt{259487} & 259487 & Yes\\
\texttt{\$259,487.00} & 25948700 & No\\
\texttt{\$2,594.87} & 259487 & Yes\\
\bottomrule
\end{tabular}
\end{center}}
\casefield{Zero witness}{For \(y^\star=0\), acceptance requires
\(0.09<\hat y<0.11\), which no parser output can satisfy. The four test
records are \artifact{tax_case_45}, \artifact{tax_case_89},
\artifact{tax_case_11}, and \artifact{tax_case_86}. One of the four training
references is also zero.
Echoing all 96 test references scores \(92/96\); this is a scorer
diagnostic and ceiling, not model accuracy.}
\casefield{Adjudication}{Under the original evaluator, zero-reference cases
cannot earn credit; the decimal parser can reject the reference amount and
accept an amount 100 times smaller.
Decimal removal was already publicly reported by \texttt{laffycat} on
September~9, 2026, in
\href{https://github.com/HazyResearch/legalbench/issues/56}{LegalBench issue~\#56}.
}
\casefield{Revision}{The scoped revision removes commas and dollar signs and parses the first number as a signed decimal amount. A missing or unparseable answer receives no credit. For a nonzero reference, the displayed inequality is applied to that parsed amount. A zero reference requires an explicitly parsed exact zero.}
\end{auditcase}

\subsubsection{A reference without its supporting rate}
\label{sec:lb-context}

LegalBench SARA entailment asks whether a proposition follows from the
supplied statute and case. In the inspected example, the target relies on a
rate omitted from the displayed statute. The omission arises in LegalBench's
published snippet; the original SARA statute retains the rate. A structural search for omitted
descendants supplies candidates for this review; it does not establish how
many entailment answers are unsupported.

\begin{auditcase}{LegalBench SARA: a rate-dependent answer without the rate}{case:lb-context}
\casesource{Entailment record \artifact{s1_b_pos}; frozen LegalBench
\href{https://github.com/HazyResearch/legalbench/tree/b46bf4ffae90524b2b72aaa30e7745fe9db64481}{code} and data~\cite{legalbench}
(\href{https://huggingface.co/datasets/nguha/legalbench/tree/daec8237410aa23e3faf4bc41ad8b3a7e1696826}{released dataset}),
and frozen SARA source.}
\casefield{Visible case and question}{Alice is a head of household in 2017
with taxable income \$97,407. The question asks whether she owes \$24,056
under section 1(b). Reference: \texttt{Entailment}.}
\casefield{Visible policy}{The introduction to section 1(b) ends with
``a tax determined in accordance with the following:''; the rates that
follow in the source are absent. The four demonstrations in each native prompt also omit these rates.}
\casefield{Full source, outside this entailment prompt}{Section 1(b)(iii)
sets \$17,544 plus 31\% of income above \$76,400, up to \$127,500.
The author program rounds its result to whole dollars.}
\caseaudit{Source comparison and arithmetic.}
\casefield{Source-supported amount}{
\[
17{,}544+0.31(97{,}407-76{,}400)=24{,}056.17.
\]
Rounding this source calculation to whole dollars gives the stored
\$24,056 amount.}
\casefield{Adjudication}{The supplied prompt lacks the rate needed to
derive that amount. A knowledge-permitted task can retain this item under
its stated contract; a supplied-policy-only decision test must restore
the relevant text or treat its information requirement separately.}
\end{auditcase}

\subsection{ConditionalQA}

ConditionalQA asks for an answer based on a policy document, together with
any conditions needed for that answer.

\subsubsection{Omitted predictions and the scoring denominator}
\label{sec:cqa-scoring}

In controlled tests (Table~\ref{tab:scorer}), the official ConditionalQA evaluator
penalizes a wrong answer but excludes an absent prediction from the aggregate
denominator. The code also divides by zero when a scoring category has no included
questions. Our revised evaluator counts every reference question,
scoring a missing ID as zero, and reports undefined category
metrics as \texttt{null}, leaving per-item scoring unchanged.

\begin{auditcase}{ConditionalQA: omitting a prediction can preserve a perfect score}{case:cqa-scorer}
\casesource{Official ConditionalQA evaluator~\cite{conditionalqa},
\href{https://github.com/haitian-sun/ConditionalQA/blob/77bd295952daf415548b3244db10880d3d55cfe0/evaluate.py}{\texttt{evaluate.py}},
\artifact{evaluate}, from the
\href{https://github.com/haitian-sun/ConditionalQA/tree/77bd295952daf415548b3244db10880d3d55cfe0}{official code and data repository};
control flow condensed.}
\caseaudit{Constructed prediction files with reference answers fixed.}
\casefield{Omission witness}{Under the original evaluator, a prediction file
echoing all 285 development reference answers receives a total exact-match
(EM) score of 1. Removing the prediction entry for \texttt{dev-0} while
echoing the other 284 leaves the score at 1. Keeping that entry but
submitting a wrong answer lowers the score to \(284/285\).}
\casefield{Control flow}{The evaluator iterates over reference questions
and initializes missing predictions' metrics to zero. But it appends
metrics and records category membership only for present predictions.
The initialized zeros therefore never enter the aggregate.}
\casefield{Derivation}{If \(I\) is the set of reference question IDs and \(M\subseteq I\) the IDs with predictions, the resulting aggregate uses \(|M|\), whereas penalizing omissions requires \(|I|\). Each omitted ID in \(I\setminus M\) is therefore excluded rather than scored as zero. A category with no appended question yields a division by zero.}
\casefield{Revision}{Accumulation moves outside the presence branch, so
missing predictions receive zero. Undefined empty-category metrics are
reported separately; per-question scoring is unchanged.
Table~\ref{tab:scorer} gives the discriminating probes.}
\casefield{Missing ID versus empty answer}{A missing ID receives zero even for an
unanswerable reference; an explicit empty answer list can receive full credit
for that same reference.}
\end{auditcase}

\begin{table}[!htbp]
\centering\small
\tablefont
\begin{tabularx}{\linewidth}{@{}Yll@{}}
\toprule
Prediction/reference configuration & Original & Revised\\
\midrule
Gold echo, all 285 dev references: total EM & 1 & 1\\
Omit dev-0; echo other 284: total EM & 1 & \(284/285\)\\
Wrong but present dev-0: total EM & \(284/285\) & \(284/285\)\\
No predictions, all 285 references & Division error & EM 0\\
143 boolean-only references, gold echo & Division error & EM 1\\
\bottomrule
\end{tabularx}
\caption{Deterministic ConditionalQA evaluator checks. ``Revised'' changes
evaluator aggregation: missing predictions count as zero and undefined
empty-category metrics are reported separately as \texttt{null}.
Per-question scoring, reference answers, and prediction inputs are unchanged.
These are controlled
diagnostic inputs, not model predictions or reported benchmark results.}
\label{tab:scorer}
\end{table}

\subsubsection{Reference answers inconsistent with the question}
\label{sec:cqa-semantics}

Case~\ref{case:cqa-companions} presents the released \artifact{dev-284}
and \artifact{dev-242} pair. Their policy link, scenario and evidence spans
are identical; their questions and stored answers differ. The supported
local target is kept separate from the unchanged-label scoring study in
Appendix~\ref{sec:cqa-study}.

In Case~\ref{case:cqa-consultation}, the policy does not support an
unconditional ``yes''. Missing case facts and uncertainty about whether
partial consultation suffices leave the replacement answer unresolved.

\begin{auditcase}{ConditionalQA: an unsupported unconditional answer}{case:cqa-consultation}
\casesource{Released \artifact{dev-111} and its policy, \emph{Making staff
redundant}; quoted spans are exact excerpts, scenario context is condensed.}
\casefield{Scenario}{The employer announces 25 redundancies,
``without bothering to run a full consultation either with our union or with
the employees as a whole.''}
\casefield{Question}{``Is it ok for the company not holding a \mbox{consultation}?''}
\casefield{Stored answer}{\texttt{yes}, with an empty condition list.}
\casefield{Document excerpts}{``You should always consult employees in a
redundancy situation.'' Redundancies without consultation ``will almost
certainly be unfair''.}
\caseaudit{}
\casefield{Finding}{This guidance does not support an unconditional yes to the
literal no-consultation question; no replacement label is established.}
\casefield{Why no replacement}{Absence of a \emph{full} consultation leaves
room for partial consultation, whose adequacy is unresolved. The collective
threshold also depends on timing and establishment facts not supplied here.}
\end{auditcase}

The audit also retains false alarms. In \mbox{\artifact{train-2032}}, the question
asks whether the applicant can request permission to appeal to the Upper
Tribunal. The policy provides such a route, subject to timing and procedural
requirements. The question concerns requesting permission, not obtaining
it or winning the appeal. The procedural qualifications alone do not
establish an error in the recorded yes, and we propose no label change.

Other source readings remain unresolved. In \artifact{train-445},
permission to add a vehicle marker does not establish whether the
authority will add it. In \artifact{train-2183}, a challenge may trigger
a hearing without determining the court's eventual finding. Neither
reading supports a simple opposite-label replacement. We also reject
the proposed yes-to-no flip in \artifact{train-1848}: requiring court
appointment does not prohibit appointing the proposed deputy.

\subsection{ShARC: identical visible inputs with different targets}
\label{sec:sharc}

ShARC asks models to answer rule-based questions or request clarification
using a rule, question, scenario, and dialogue history. These appear in the
model-visible fields \texttt{snippet}, \texttt{question}, \texttt{scenario},
and \texttt{history}; the auxiliary \texttt{evidence} field is not shown to
the model~\cite{sharc}.\footnote{Input-field documentation: \url{https://sharc-data.github.io/data.html}. Archive: \url{https://sharc-data.github.io/data/sharc1-official.zip}.} The inspected archive contains answers for all three splits. We group it by
the exact ordered tuple of these fields within each split, preserving text and
list order. The check finds one train group with three records and one dev
group with two records whose targets differ. The train group has two yes
targets and one clarification target. No such group is found in the 8,276
test records under exact matching; semantically equivalent but differently
worded inputs are outside this check.

Together these are five records in two incompatible-target groups across
different splits. A
deterministic response cannot match every single-reference target in either
group. In the dev group, a negative answer to the status question appears
only in one record's hidden evidence
(Case~\ref{case:sharc}). Repair requires adjudicating the visible case and its
clarification requirements, whereas exposing hidden evidence changes the input
and may reveal the answer.

\begin{auditcase}{ShARC: two targets for the same visible input}{case:sharc}
\casesource{Released dev records; visible fields condensed, target wording
quoted exactly.}
\casefield{Records}{Dev source indices 960 and 2106.}
\casefield{Shared question and history}{Both ask ``Do I qualify for this
benefit program?'' and have an empty dialogue history.}
\casefield{Visible scenario and policy (condensed)}{The speaker lives in
Minnesota and plans to study nursing. The policy requires Native
American/American Indian status, acceptance or enrollment in an accredited
health-care degree program, and enrollment of the user or a family member
in a federally recognized American Indian tribe or Alaska Native village.}
\casefield{Targets}{Index 960: ``Are you a Native American/American Indian?''
Index 2106: ``No''.}
\casefield{Hidden annotation}{Index 2106's auxiliary \texttt{evidence} supplies
a negative answer to the status question. This field is not shown to the model.}
\caseaudit{}
\casefield{Grouping}{The full ordered visible tuples (\texttt{snippet},
\texttt{question}, \texttt{scenario}, \texttt{history}) of the two records are
identical under exact matching.}
\casefield{Finding}{The visible records support no deterministic way to
reproduce both single-reference targets. Their four visible fields are identical; the negative status answer appears only in index 2106's
hidden evidence.}
\casefield{Adjudication}{One incompatible-target group with unresolved
clarification requirements; neither target is selected as a replacement.
Adding the hidden evidence to the
input would define a different task.}
\end{auditcase}

\subsection{PolicyGuardBench: conflicting targets for identical inputs}
\label{sec:pgb}

The original and strict releases contain the same labeled multiset of
59,997 records. The strict release repartitions these records, reducing
shared train--test trajectories from 592 to zero. Grouping by complete
instructions and inputs finds 279 conflicting binary-label groups
involving 780 records in each release; every conflicting group is
preserved. Case~\ref{case:pgb-conflict} illustrates one such conflict.

\begin{auditcase}{PolicyGuardBench: a split repair preserves target conflicts}{case:pgb-conflict}
\casesource{Original and strict PolicyGuardBench
releases~\cite{policyguardbench}; exact complete-input \mbox{comparison}.}
\casefield{Task}{Classify a supplied web-agent trajectory summary as
\texttt{violation} or \texttt{no\_violation} under a policy.}
\casefield{Selected records}{Original train index 8 and test index 2937,
using zero-based source positions.}
\casefield{Shared policy}{Require user confirmation before clicking the
Products link when the risk level is high.}
\casefield{Shared input}{The classification instruction, all 18 action
lines, and the \texttt{shopping\_admin}/\texttt{dashboard} metadata are
identical, including the source's final unknown action.}
\casefield{Targets}{Train index 8: \texttt{violation}.
Test index 2937: \texttt{no\_violation}. In the strict release, the same
input appears at train indices 6 and 41140 with these opposing labels,
respectively.}
\casefield{Second input, within strict test}{Zero-based test indices 7294
and 7962 have labels \texttt{violation} and \texttt{no\_violation}.
Their common policy prohibits access to admin-system account pages when
risk is high. Their two action lines are condensed here as ``click
REPORTS; unknown'', with \texttt{shopping\_admin}/\texttt{dashboard}
metadata. The complete instruction and input strings are identical,
including both action lines and all metadata; only the output label
differs. This pair belongs to one of the 60 conflicting groups within
the strict test split.}
\caseaudit{Exhaustive grouping in both releases.}
\casefield{Adjudication}{A deterministic classifier cannot match both
targets for an identical supplied input. Lossy source summaries and label
noise remain competing explanations; neither target is chosen as correct.
The resolved split-overlap issue is not a current strict-release defect.
Conflict-group exclusion is a possible population revision; replacement
targets remain unadjudicated. BAR records the conflicts and consistency
bounds without releasing a corrected label set. These conflicts are not assigned to adopter scores whose
evaluated version is unknown.}
\end{auditcase}

The strict release still contains 60 conflicting complete-input groups
in its test split, involving 143 of 11,545 rows. Summing the less frequent
label count in each group gives 61: any deterministic prediction based only
on the complete visible input must disagree with at least this many test
labels. This is a consistency bound, not 61 adjudicated label errors or an
observed model result. The inconsistency therefore remains within the
held-out evaluation population.

The full consistency counts are shown in Table~\ref{tab:pgb-edits}.
For each complete-input group, the minimum change count is the smaller
of its two label counts; summing over groups yields the bound. Pooled
grouping combines train and test before counting, so its minimum need not
equal the sum of the two within-split minima.

\begin{table*}[t]
\centering\small
\begin{tabular}{@{}llrrrr@{}}
\toprule
Release & Population & Rows & Conflicting groups & Involved rows & Minimum edits\\
\midrule
Original & Train & 47,997 & 196 & 531 & 220\\
 & Test & 12,000 & 22 & 49 & 22\\
 & Pooled & 59,997 & 279 & 780 & 324\\
\addlinespace[3pt]
Strict & Train & 48,452 & 219 & 637 & 263\\
 & Test & 11,545 & 60 & 143 & 61\\
 & Pooled & 59,997 & 279 & 780 & 324\\
\bottomrule
\end{tabular}
\caption{PolicyGuardBench minimum label edits for identical-input
consistency. Groups are identical complete inputs with both targets;
involved rows are their members. Pooled combines train and test within
each release. Strict repartitions train/test while preserving the same labeled
multiset and its target conflicts. Minimum edits are lower bounds for
consistency, not counts of adjudicated label errors.}
\label{tab:pgb-edits}
\end{table*}

\subsection{Workflow evaluators: coverage of required decisions}
\label{sec:workflow-grading}

These checks test whether evaluator acceptance covers the requested
decisions. Case~\ref{case:tac-budget} executes the official scoring chain to
test whether acceptance requires a reported budget decision;
Case~\ref{case:tac-users} tests whether accepted records cover distinct users.
For the main-text WorkArena++
Case~\ref{case:workarena-gap}, this appendix supplies the source branch
and constructed-state controls for the scheduling limit.

\begin{auditcase}{TheAgentCompany: totals earn credit without a budget decision}{case:tac-budget}
\casesource{TheAgentCompany~\cite{theagentcompany} task
\artifact{admin-check-employees-budget-and-reply}; task instruction,
checkpoint 3, and final-checkpoint bonus.}
\casefield{Policy and task}{Each of three departments has a \$10,000 budget.
Obtain equipment requests and prices, then tell each manager both the total
cost and whether the budget is sufficient.}
\casefield{Reference and grader}{The evaluator searches each manager chat for the corresponding expected
total. It never tests a sufficiency statement.}
\caseaudit{Controlled replies stored in real manager chats, scored through
the unchanged official chat reader, checkpoints, completion bonus, and
final score conversion.}
\casefield{Matched conditions}{The replies use the literal total strings
\texttt{11917}, \texttt{5001}, and \texttt{18780}, which the evaluator
matches as substrings. A reports these totals and
consistent budget judgments; B omits the judgments; C reverses all three
judgments. D changes only the first total in A to \$11,918. Each condition
starts from the same database snapshot, with one seeded message already
present in each chat, and uses the same constructed trajectory input.
The seeded messages satisfy the first checkpoint independently of the
final replies.}
\smallskip
{\centering
\setlength{\tabcolsep}{4pt}
\begin{tabularx}{\linewidth}{@{}Ycrr@{}}
\toprule
Replies & Checkpoints & Points & \shortstack[r]{Converted\\score}\\
\midrule
A: consistent judgments & \(1,1,2\) & \(4/4\) & \(1.00\)\\
B: judgments omitted & \(1,1,2\) & \(4/4\) & \(1.00\)\\
C: judgments reversed & \(1,1,2\) & \(4/4\) & \(1.00\)\\
D: one total changed & \(1,1,0\) & \(2/4\) & \(0.25\)\\
\bottomrule
\end{tabularx}\par}
\smallskip
Points and converted scores are official outputs; the final score is not
the raw point fraction. The unchanged earlier checkpoints pass in all
conditions, so this check does not separately identify the bonus's effect.
\casefield{Adjudication}{The evaluator leaves the required budget decisions
unchecked in these constructed states. The totals come from the evaluator;
prices and amount references are not independently certified. No model
produced the replies, and the trajectory contains a constructed directory
marker rather than a recorded browser visit. The check establishes scoring
behavior, not its frequency in agent episodes.}
\casefield{Prior report and repair status}{
\href{https://github.com/TheAgentCompany/TheAgentCompany/pull/888}{TheAgentCompany PR~\#888}
previously exposed the total/sufficiency mismatch. It added the total-cost
requirement to the task and left the evaluator unchanged.
A revised evaluator that also checks budget sufficiency would require
separate validation.}
\end{auditcase}

\begin{auditcase}{TheAgentCompany: five rows do not establish coverage of five users}{case:tac-users}
\casesource{TheAgentCompany's batch password-update task~\cite{theagentcompany},
initial file, and \mbox{evaluator}.}
\casefield{Task}{Contact only users whose passwords are invalid and leave
correct password records for every team member. The initial file contains
five users, two with invalid passwords and three with valid ones.}
\casefield{File checkpoint}{The file check counts rows matching stored
username--password pairs and accepts exactly five matches. It does not
check the set of users represented.}
\caseaudit{Constructed files and static predicate analysis.}
\casefield{Witness and controls}{Five copies of one initially visible valid
row satisfy the file check while omitting four users. Five distinct correct
rows also pass; four matching rows and five wrong rows fail.}
\casefield{Contact checkpoint and scope}{The file checkpoint is worth two
of four points. Full credit also requires the specified contact
state: the two initially invalid users contacted and none of the other
three. No live episode was run. The supported defect is counting records
instead of checking coverage; a revision must verify the required user set.}
\end{auditcase}

\paragraph{Source branch and controls for Case~\ref{case:workarena-gap}.}
The selected scheduling class sets \texttt{goal\_type="tight priority"}.
The validator uses \texttt{1 if goal\_type == "tight" else 24} hours, so
this class receives the looser limit. The Company Protocols manual and the
L2 prompt require a gap of at most one hour; the L3 task refers to the tight schedule
under that manual.

The supplied state contains two risk-3 requests of equal impact, each
requiring two days. Both fit the generated time window without overlap.
The selected class accepts the two-hour gap and a one-hour-gap control.
The separate plain-tight control instead uses two risk-4 requests requiring
one day each and rejects a two-hour gap. The controls share the gap duration,
not the entire state; their risk and required durations follow their
respective task modes. Overlap, a gap above 24 hours, a wrong duration,
and a missing date fail. L3 additionally requires the stated completed
private-ticket state. These are native-method diagnostics on supplied states,
not live ServiceNow episodes.

The wider threshold establishes false acceptance in this class's spacing
check and cannot by itself create false rejection. A one-hour check for
this class remains a scoped revision candidate. Sibling task bindings and actual adopter effects
remain unverified.

\subsection{Historical repairs and boundary cases}
\label{sec:historical-controls}

\subsubsection{\texorpdfstring{$\tau$}{tau} banking: a repaired state comparison}
\label{sec:tau}
The later \texttt{banking\_knowledge} extension provides a historical
false-rejection control outside the nine-benchmark audit.
Before v1.0.1, an additional valid read could change the tool log in the
compared database without changing business records. Upstream
\href{https://github.com/sierra-research/tau2-bench/pull/329}{PR~\#329}
and the \href{https://github.com/sierra-research/tau2-bench/blob/4ce7c0397c1eb65c9bbe59aeacfe1ca44a1cd699/CHANGELOG.md}{v1.0.1 changelog}
document the repair.

Replaying task~075's reference actions and then an account-list lookup
fails the historical state-equality check and passes in v1.0.1.
Omitting its account-opening call or opening an unrequested account
fails in both versions. For task~070, omitting its required lookup fails
in both versions, while repeating that lookup passes in both.
The task definitions are unchanged; the repair removes the extra-read
penalty while retaining required actions and business-state checks.
These database-component controls establish no model success rate and
no persisting defect in the fixed version.

A separate change in the same release increases reference refund
\texttt{074\_10} from \$8 to \$14.50 to account for a free ATM allowance.
That historical reference correction does not cause the logging results.

\subsubsection{SOP-Bench: discretion and single-reference evaluation}
\label{sec:amend-discretion}

As a boundary example beyond the nine-benchmark audit, SOP-Bench traffic
spoofing permits investigator discretion~\cite{sopbench}. Risk level is a declared input. The released
200 cases contain 92 High/Closure, 50 Medium/Suspension, 39 Medium/Warning, and
19 Low/No Action pairs. The current policy makes suspension the medium-risk
default but allows a warning based on investigator assessment (see the
discretion replies to
\href{https://github.com/amazon-science/SOP-Bench/issues/5}{SOP-Bench issue~\#5}).

The author reply treats default-based and contextual decisions as informative
interpretations. Our terminal-rule replay confirms that the 39 stored warnings
differ from the medium-risk default, while the policy permits contextual
warnings; their case-level justification remains unresolved. A valid-response-set
evaluation or a more explicit decision boundary would measure a different
contract from matching a single recorded choice.

\paragraph{An amended decision rule.}
SOP-Bench's referral-abuse
\href{https://github.com/amazon-science/SOP-Bench/issues/4}{issue~\#4}
reported nine strict disagreements and two cross-action ties under the
earlier rule, which selected the highest eligible category score.
The subsequent
\href{https://github.com/amazon-science/SOP-Bench/commit/1fdeba5949c8ef07f0c9a7e3a3b263f0999bd292}{policy amendment}
prioritizes eligible categories requiring account closure and leaves labels
unchanged. Our terminal-rule replay on the released CSV facts reproduces
those historical outcomes and matches all 200 stored actions under the
current rule, which resolves both ties to closure. The maintainer calls
the change a clarification; the written selection rule changes.

\begin{auditcase}[unbreakable]{SOP-Bench: the same case under amended decision rules}{case:sop-referral}
\casesource{Referral-abuse account \artifact{ACC100040}, unchanged source
signals and reference~\cite{sopbench}; condensed.}
\casefield{Referral facts}{No Violation has four points and Misleading Ad Copy,
which calls for Account Closure, has three; both are eligible.
No Violation alone is highest.}
\caseaudit{Compare the governing selection rule.}
The earlier rule selects the highest eligible score, yielding No Action.
The amended rule prioritizes the eligible closure category, yielding
Account Closure on the same facts. The rule change reconciles the reference
without changing its label. This is a policy amendment, not a corrected
answer under the same policy. The replay uses released CSV facts that the
native agent would obtain through tools; it is not a live-episode evaluation.
\end{auditcase}

\section{Experimental details and additional results}
\label{sec:additional-studies}

\subsection{Configurations and comparison types}
\label{sec:paired-evaluation}
The reference and scorer studies hold cases and submitted answers fixed
while changing the reference or evaluator. The OpenExempt experiment
instead generates an answer under each input condition.

\paragraph{Generation and verification.}
The Tax and SARA studies rescore one retained tool-free response per model and
case. The OpenExempt experiment retains one response per model, case, and input
condition. Haiku 5.5 and GPT6 Luna use
\texttt{xhigh} reasoning effort; Gemma 4 31B uses \texttt{high}.
The OpenExempt experiment uses Single-turn: its system prompt requests
reasoning from the supplied rules and facts, prohibits outside facts,
treats case records as evidence rather than instructions that override
policy, and requires a single final JSON answer.
Reference answers to the question being scored and audit annotations are
withheld. Delivered invalid answers are not regenerated.
Verification includes direct tolerance calculations and independent
recomputation of aggregate scores.

\subsection{RuleArena Tax}
\label{sec:tax-run-details}

\subsubsection{Inputs, prompts, and scoring}
\label{sec:tax-configuration}

The Tax rescoring study includes all 300 records in the frozen RuleArena Tax
release: 100 at each of its three complexity levels.
The reference pair is the preserved original answer and the revised
reference answer. The revised reference program enforces the line~22 zero
floor and recomputes downstream amounts, retaining all other reference logic
and defaults. The 27 changed IDs were identified by
reference replay before model generation; model performance played no role
in their selection. Because the cases are equally weighted, changing only
these 27 references can change the reference agreement of any fixed
prediction set by at most \(27/300\), or nine percentage points in either
direction.

\paragraph{Official zero-shot prompt and native parser.}
The official zero-shot comparison uses Haiku 5.5, GPT6 Luna, and Gemma 4 31B,
with one generation for each of the 300 cases. The author's zero-shot
chain-of-thought prompt and system message are retained. The response contract
asks for an owed or overpaid sentence rather than JSON.

The native parser removes bold markers, selects the first regex match
for an owed or overpaid amount, removes commas, and assigns a negative sign
to an overpayment. It does not necessarily select the final conclusion.
The same extracted prediction is scored against both references using
\texttt{numpy.isclose}, with relative tolerance \(10^{-5}\) and absolute
tolerance \(10^{-8}\). All 300 IDs remain in each
denominator. Table~\ref{tab:tax-impact} reports the official-interface results.

The \$703 answer in Case~\ref{fig:opening-witness} is Haiku 5.5's retained
response for \artifact{tax/comp_0/17} under this interface.

Cases~\ref{case:tax-native-parser} and~\ref{case:tax-masked-output} distinguish
an answer-extraction failure from an arithmetic error hidden by final-answer
agreement. They use Haiku 5.5 and Gemma 4 31B responses,
respectively, under the official zero-shot interface.

\begin{auditcase}{Tax: the native parser selects an earlier amount}{case:tax-native-parser}
\casesource{RuleArena Tax's native parser and an observed response for \artifact{tax/comp_0/34}; response excerpts quoted.}
\casefield{Task and sign convention}{Report the final amount owed or
overpaid; refunds are represented as negative amounts.}
\casefield{Earlier sentence}{``The total tax owed is \$0 (the total tax is
\$763, which is fully covered by the \$17,150 in payments).''}
\casefield{Final sentence}{``The total tax overpaid is \$16,387.''}
\casefield{Reference pair}{The zero-floor correction changes the reference
from \(-16{,}552\) to \(-16{,}387\).}
\caseaudit{Native extraction and scope.}
\casefield{Extraction}{The parser extracts \(0\) from the earlier sentence,
not the \(-16{,}387\) represented by the final refund statement.
The extracted \(0\) is rejected against both references, although the
final sentence matches the revised reference.}
\end{auditcase}

\begin{auditcase}{Tax: final agreement despite an arithmetic error}{case:tax-masked-output}
\casesource{Observed response and supplied education-credit worksheet for
\artifact{tax/comp_1/42}; the addition is reproduced from the response.}
\casefield{Model calculation}{The response adds nonrefundable tax credits as
\(141+1{,}918+731+1{,}459+3{,}592=8{,}841\). It subtracts this amount from
a tax subtotal of \(6{,}382\) and applies a zero floor. It then subtracts
payments from other taxes, giving \(1{,}077-7{,}563=-6{,}486\), which matches
the reference after the zero-floor correction.}
\caseaudit{Check the displayed addition and the zero floor.}
\casefield{Check}{The five displayed credit amounts sum to \(7{,}841\), not
\(8{,}841\). Either sum exceeds \(6{,}382\), so either produces zero after
the floor. Agreement on the final amount therefore conceals this arithmetic
error.}
\casefield{Conditional worksheet check}{The education-credit worksheet
limits that credit to the tax subtotal minus specified other credits.
The response includes Schedule 3 lines 6i and 6j, although the worksheet
lists only lines 1, 2, 6d, and 6l. Retaining the reference's zero default
for the missing line 6l, the listed credits sum to \$2,375 rather than
\$5,651. The corrected limit is \(6{,}382-2{,}375=4{,}007\), so the
response's pre-limit education credit of \$1,990 is retained instead of
being reduced to \$731. With that correction, total credits still exceed
the \$6,382 tax subtotal, leaving line 22 at zero and the final amount
unchanged.}
\end{auditcase}

\subsubsection{Tax output traces}
\label{sec:tax-output-attribution}

A model-assisted pass reviews the response behind every changed judgment
under the official zero-shot interface: 17 for Haiku 5.5, 20 for GPT6 Luna,
and 25 for Gemma 4 31B (Table~\ref{tab:tax-native-traces}). Scores were
visible during review.
An explicit floor states a \texttt{max} operation, explains that excess
credits set the subtotal to zero, or shows a negative subtotal explicitly
truncated to zero with an arrow. A zero subtotal alone does not establish
that reasoning. Each attribution is checked against the response text.
Of the 61 gains, 48 show an explicit floor and 13 a zero subtotal without
the stated rule. Case~\ref{case:tax-compensation} traces Haiku's sole lost
acceptance.

\begin{table}[!htbp]
\centering
\small
\begin{tabular}{@{}lrrrr@{}}
\toprule
 & \multicolumn{2}{c}{Gains} & \multicolumn{2}{c}{Losses}\\
\cmidrule(lr){2-3}\cmidrule(lr){4-5}
Model & \shortstack[r]{Explicit\\floor} & \shortstack[r]{Zero\\subtotal\\only} &
\shortstack[r]{Explicit\\floor} & \shortstack[r]{Zero\\subtotal\\only}\\
\midrule
Haiku 5.5 & 16 & 0 & 0 & 1\\
GPT6 Luna & 10 & 10 & 0 & 0\\
Gemma 4 31B & 22 & 3 & 0 & 0\\
\bottomrule
\end{tabular}
\caption{Observable floor evidence behind all 62 changed judgments under the
official zero-shot Tax interface. An explicit floor states the operation;
a zero subtotal alone does not.}
\label{tab:tax-native-traces}
\end{table}

\begin{auditcase}[unbreakable]{Tax: a lost acceptance from compensating calculations}{case:tax-compensation}
\casesource{Haiku's observed official-interface response for
\artifact{tax/comp_1/69} and its reference pair.}
\casefield{Task}{Report tax owed or a refund; negative amounts denote refunds.}
The unchanged answer \(-10{,}771\)
matches the original reference but not the revised amount \(-8{,}528\).
The response limits the child credit to
\(2{,}257\) using a worksheet that subtracts all Schedule~3 credits, so the
line~22 subtotal is \(9{,}938-9{,}938=0\). It then refunds the remaining \(2{,}243\) as additional child tax
credit.
\caseaudit{Compare the response and reference calculations.}
No negative subtotal requires a floor in this response.
The extra \$2,243 refund equals the difference between the original and
revised reference refunds: \(10{,}771-8{,}528=2{,}243\).
It reproduces the original reference's negative line-22 effect through a
different calculation, so the final amount matches the original reference.
The lost acceptance reflects
compensating calculations, not a demonstrated failure to apply the floor.
\end{auditcase}

Conversely, Gemma's revised match on \artifact{tax/comp_1/42}
conceals an addition error (Case~\ref{case:tax-masked-output}) and a
worksheet discrepancy under the retained missing-field default.
Accepted answers that reach the
low-earnings self-employment branch or the Schedule~3 zero default do not
resolve those dependencies.

\subsection{LegalBench SARA: numeric scorer effects}
\label{sec:sara-study}
The study uses all 96 numeric TEST rows, excluding the four TRAIN rows,
with unchanged statutes, case descriptions, questions, and references.
Case~\ref{case:lb-scorer} defines the original and revised scoring rules.

\paragraph{Prompt and scoring.}
Each model receives the author's unchanged zero-shot template as its sole
user message, with no added system task instruction, and produces one
visible response in a tool-free call. The template asks to state the amount
first and imposes no integer-only format.
The response in Case~\ref{case:lb-summary} is Gemma 4 31B's retained answer
for \texttt{tax\_case\_25}.
Without an answer-extraction follow-up, the entire visible text is passed
unchanged to three evaluator variants: the original scorer, the parsing-only
revision, and the revision of parsing and zero handling. The parsing-only
variant reads the first signed decimal amount, without consulting the
reference, but keeps the released formula. For a zero reference that
formula still rejects an exact zero and would accept \(0.10\); it is a
diagnostic, not a usable repair. Adding zero handling enforces the exact-zero rule.

The first amount found by the revised parser is
a decimal in 82 Haiku, 76 Luna, and 78 Gemma responses and an integer in the rest; none has
an explicit sign or lacks a parseable amount. Parsing accounts for 229 of
the 241 changed judgments and the exact-zero rule for the remaining 12.
No accepted answer becomes rejected.

\subsection{ConditionalQA: aggregation effects and complete-file controls}
\label{sec:cqa-study}
We use all 285 public dev questions, supplying the full associated policy
document, question, and scenario. Reference answers are withheld. Our
Single-turn interface requests a JSON-wrapped native answer item, including
answer text and any conditions. Failed submissions become missing
prediction IDs; an explicit empty answer list remains a valid prediction
of unanswerability.

Both evaluators receive the same decoded predictions and original
reference labels. Neither this rescoring nor the complete-file controls
below apply the local semantic label revision in
Case~\ref{case:cqa-companions}. The original averages over submitted questions. Our revision
assigns missing predictions zero and includes every reference question.
This is an explicit coverage rule; rejecting incomplete files would be
another option. Answer matching and condition scoring are unchanged.

EM compares normalized answer text exactly; F1 measures token overlap.
The condition-aware variants also multiply answer credit by F1 over the
associated condition sets. An explicitly empty answer list can receive
full credit for an unanswerable question; a missing prediction ID cannot
under the full-coverage rule.
The 285 questions comprise 143 yes/no, 128 extractive, and 14 unanswerable
questions; the 63 conditional questions fall within the first two groups.

\begin{table*}[t]
\centering\small
\begin{tabular}{@{}lrrrrr@{}}
\toprule
Model & Denominator & EM & EM+cond. & F1 & F1+cond.\\
\midrule
Haiku 5.5 & \(285\to285\) & \(58.63\to58.63\) & \(34.64\to34.64\) & \(65.81\to65.81\) & \(40.22\to40.22\)\\
GPT6 Luna & \(261\to285\) & \(52.34\to47.93\) & \(38.56\to35.31\) & \(61.97\to56.75\) & \(46.75\to42.81\)\\
Gemma 4 31B & \(232\to285\) & \(58.02\to47.23\) & \(42.45\to34.55\) & \(66.22\to53.91\) & \(48.96\to39.86\)\\
\bottomrule
\end{tabular}
\caption{ConditionalQA fixed-output aggregation effects
(original \(\to\) revised; metrics in percent).
All models attempt the same 285 questions using Single-turn.
EM measures normalized answer matching; F1 measures token overlap.
The condition-aware variants also score the predicted condition sets.
The revision changes denominator membership, not per-question scores.
Submission failures belong to our interface, not to the reference labels.}
\label{tab:cqa-impact}
\end{table*}

Table~\ref{tab:cqa-impact} shows that excluding missing predictions inflates
Luna's and Gemma's scores relative to the all-question rule. Gemma has the
highest answer F1 on the submitted subset and the lowest after including
missing predictions. This change concerns these saved outputs and their
submission coverage, not a stable ranking of reasoning ability. Haiku
submits an answer for every question and its scores are unchanged.

The original category denominators (yes/no, extractive, conditional)
are 132/116/56 for Luna and 118/101/48 for Gemma; the revision restores
143/128/63. Haiku's category denominators are unchanged. Valid empty-answer
predictions number 9, 17, and 16 for Haiku, Luna, and Gemma, respectively.
Missing or malformed predictions are not reconstructed from unsubmitted
fragments.

To check whether full-coverage aggregation changes scores when no answers
are omitted, we rescored eight unchanged ConditionalQA prediction files published
with Code Prompting~\cite{codeprompt}\footnote{Official experiment repository: \url{https://github.com/UKPLab/emnlp2024-code-prompting}.}: TextPrompt and CodePrompt with
GPT-3.5 (three source runs per method) and Mistral-7B (one per method).
Every file covers all 285 dev IDs. All four metrics---answer exact match,
token-overlap F1, and their condition-aware versions---and all category
denominators are identical under the original and revised evaluators.
These files repeat the same questions under historical oracle-context or
code-translation inputs. They test the evaluator revision, not a new model
configuration.

\subsection{OpenExempt: effects of supplied intermediate answers}
\label{sec:oe-evidence}\label{sec:oe-step-design}

This experiment tests whether amounts presented as already solved affect a
later decision. Selection screens all 100 OE4 records and includes the 69
whose released solved steps contain federal amounts. The other 31 are
excluded before generation. The experiment asks for the minimum total
non-exempt value under the federal election, rather than the source task's
choice of jurisdiction and exact allocation schedule.
All three conditions explicitly stipulate that the debtors are spouses
in a joint case, both elect federal exemptions, their equal half interests
are unencumbered, the listed assets form the complete estate, and no other
claims consume the allowances. Eligibility and asset categories are given
for each spouse. Each asset entry, including a set or bundle, is stipulated
to be one indivisible item for per-item limits. These added premises make
the optimization question determinate in the experiment; they are
not attributed to the native case narratives.

The supplied-step conditions identify each amount as the combined maximum
for one asset--provision pair before other pairs consume shared allowances.
Released steps supply the released federal amounts; Revised steps supply
per-pair maxima computed under the stated case assumptions. The third
condition omits the amounts while retaining the same eligibility
information. Policy, question, and other case facts are identical across
conditions. The revised supplied steps change amounts in 65 cases; four unchanged
tables provide controls (Table~\ref{tab:oe-step-controls}).

\paragraph{Independent target calculation.}
One calculation enumerates the permitted vehicle choices and solves a
linear allocation problem under per-item and shared caps. A second uses
decimal arithmetic to maximize the category-specific exemptions for one
spouse, apply the remaining wildcard allowance to the residual, and double
the result for equal ownership. The wildcard is an allowance usable on any
asset: under the supplied federal rules, each spouse has a base \$1,475
plus up to \$13,950 of the unused \$27,900 homestead allowance. Using the
homestead exemption can therefore reduce the addition to the wildcard. The two
calculations agree on all 69 targets. They do not use the native terminal
labels as gold. A finite nonnegative prediction is accepted within
\$0.005 of this target; an invalid answer remains in the denominator.

\paragraph{Sensitivity and controls.}
Before generation, a separate optimization treats the released pair
amounts as additional ceilings. It changes the minimum in 19 of the 65
cases with changed tables. Across all 69 cases, 50 minima are unchanged.
The revised ceilings preserve the independent optimum in all 69 cases.
There are 20 positive-residual and
49 zero-residual targets.

\begin{table}[!htbp]
\centering\small
\begin{tabularx}{\linewidth}{@{}Yccc@{}}
\toprule
Control (cases) & Haiku & Luna & Gemma\\
\midrule
Unchanged tables (4) & 4/4/4 & 4/4/4 & 4/4/4\\
Positive residual (20) & 4/19/20 & 17/19/20 & 16/16/20\\
Zero residual (49) & 46/49/49 & 49/49/49 & 49/48/49\\
Ceiling matches (19) & 5 & 3 & 0\\
\bottomrule
\end{tabularx}
\caption{OpenExempt Single-turn input controls. Triples give accepted counts for
Released steps / Revised steps / No supplied steps, each generated
separately. The residual rows partition the 69 cases; the other rows overlap
these groups. Only supplied intermediate amounts vary; revised amounts are
per-pair maxima under the shared case assumptions. Ceiling matches count
released-step outputs matching the optimum under the released ceilings on
the 19 sensitive cases; they do not attribute every error to those ceilings.}
\label{tab:oe-step-controls}
\end{table}

\paragraph{Relation to native findings.}
Within these 69 adapted cases, 65 supplied tables change, and treating the
released amounts as ceilings changes the target in 19. The native audit's
38 conflicting pairs in 33 of the 100 OE4 records concern the released
allocations, not these experimental targets.
The native allocator diagnostic finds a positive second federal claim in
36 of the 69 experimental cases, including all 19 sensitive cases.
Separately, 14 of the 69 cases, including 13 of the sensitive cases, have
a released final allocation exceeding its supplied maximum. These checks
overlap and inspect different components. The 65 revised tables are
computed under the added experimental premises; they are not 65
independently adjudicated native errors. Nor do these overlaps attribute
each change in the experimental target to the one-claim defect alone.

\paragraph{New rejections under revised steps.}
Gemma's three new rejections contain valid JSON amounts. In records 24 and
31, the responses omit the unused-homestead addition to the any-property
allowance, leaving too much value unprotected. In record 51, the response
leaves part of that allowance unused while the homestead retains
unprotected value. These numerical errors account for its three losses;
the two gains give the net decrease from 65 to 64 accepted responses.
Luna's loss in record 47 instead states the target amount in prose but
omits the required JSON object.

Case~\ref{case:oe-step-response} compares Haiku 5.5 Single-turn responses
with released, recalculated, or no supplied intermediate amounts. The policy,
question, and other case facts remain fixed.

\begin{auditcase}[unbreakable,arc=.5mm,halign title=flush left]{OpenExempt: a supplied amount changes the calculation}{case:oe-step-response}
\casesource{Haiku 5.5 responses in our federal-exemption experiment, based
on OE4 record 23.}
\casefield{Experimental premises and question}{This experiment specifies a
joint case in which both spouses elect federal exemptions, own equal shares
of assets free of secured debt, and list all their assets. What is the minimum
total asset value left unprotected from creditors?}
\casefield{Available exemptions}{A \$54,990 car qualifies for \$4,450 per
spouse, or \$8,900 combined. Other assets are fully protected by separate
provisions, leaving the entire \$30,850 any-property allowance available
for the car.}
\caseaudit{With Released steps (the released intermediate amounts), the
response treats \$4,450 as the combined vehicle limit and calculates:
\[
54{,}990-4{,}450-30{,}850=\mathbf{19{,}690}.
\]
The separately generated Revised response (with recalculated amounts) and
None response (without supplied amounts) use \$8,900 and each report
\textbf{\$15,240}.}
\end{auditcase}

Each model produces one answer per condition. The observed transitions
therefore describe these generations, not a repeated-run estimate. The
released-step condition deliberately retains disagreement between the
policy-based target and supplied amounts. Its score measures success
against that target, including whether the model overcomes the conflicting
intermediate information.

\end{document}